\documentclass[10pt,journal,compsoc]{IEEEtran}

\usepackage{tikz}
\usetikzlibrary{arrows,decorations.markings,fadings,shadows,shapes.arrows}

\usepackage{url}
\usepackage{array}
\usepackage{xparse}
\usepackage{graphicx}
\usepackage{subfigure}
\usepackage{amssymb}
\usepackage[linesnumbered,lined,boxed]{algorithm2e}

\usepackage{cite}

\begin{document}

\title{Measuring economic activity from space: a case study using flying airplanes and COVID-19}

\author{ 
         Maur\'icio~Pamplona~Segundo,~\IEEEmembership{Member,~IEEE,}
         Allan~Pinto,~\IEEEmembership{Member,~IEEE,}
         Rodrigo~Minetto,~\IEEEmembership{Member,~IEEE,}
         Ricardo~da~Silva~Torres,~\IEEEmembership{Member,~IEEE,}
         and 
         Sudeep~Sarkar,~\IEEEmembership{Fellow,~IEEE}
\IEEEcompsocitemizethanks{
\IEEEcompsocthanksitem M. Pamplona Segundo and S. Sarkar are with Department of Computer Science and Engineering, University of South Florida (USF), Tampa, FL, USA. E-mail: \{mauriciop,sarkar\}@usf.edu
\IEEEcompsocthanksitem A. Pinto is with Universidade Estadual de Campinas (UNICAMP), Brazil. E-mail: allan.pinto@ic.unicamp.br
\IEEEcompsocthanksitem R. Minetto is with Universidade Tecnol\'{o}gica Federal do Paran\'{a} (UTFPR), Brazil. E-mail: rminetto@utfpr.edu.br
\IEEEcompsocthanksitem R. da Silva Torres is with Norwegian University of Science and Technology, Department of ICT and Natural Sciences, Alesund, Norway. E-mail: ricardo.torres@ntnu.no
}
}

\maketitle

This work introduces a novel solution to measure economic activity through remote sensing for a wide range of spatial areas. We hypothesized that disturbances in human behavior caused by major life-changing events leave signatures in satellite imagery that allows devising relevant image-based indicators to estimate their impacts and support decision-makers. We present a case study for the COVID-19 coronavirus outbreak, which imposed severe mobility restrictions and caused worldwide disruptions, using flying airplane detection around the 30 busiest airports in Europe to quantify and analyze the lockdown's effects and post-lockdown recovery.  Our solution won the Rapid Action Coronavirus Earth observation (RACE) upscaling challenge, sponsored by the European Space Agency and the European Commission, and now integrates the RACE dashboard. This platform combines satellite data and artificial intelligence to promote a progressive and safe reopening of essential activities. Code and CNN models are available at \url{https://github.com/maups/covid19-custom-script-contest}.


\section{INTRODUCTION}\label{sec:introduction}

Our planet is experiencing an increase in disasters\cite{UNDRR2020} and disease outbreaks\cite{Smith2014} over the past decades. Therefore, developing methods and tools to provide meaningful information for assertive decision-making during emergencies, thus implementing safety and welfare measures, is of paramount importance. Such a demand has led the research community to focus on finding indicators to support the different phases of emergency management\cite{Raymond2020,Eyre2020,Simon2015,Nature2020,Minetto2020}. One of the main challenges in designing an indicator, especially when its coverage area is beyond the country level, is collecting data that helps analyze a phenomenon and improves our understanding of its causes and symptoms. Depending on the required information ({\it e.g.}, data from cell phones\cite{Chang2021}, social media, press releases), different factors spoil this process, including access limitations, varying technologies, lack of infrastructure, sovereignty restrictions, and language mismatch, to list a few. We hypothesize that disruptions of this magnitude impact social behaviors and leave signatures in satellite imagery that can be automatically detected and quantified. And unlike other data sources, remote sensing stands out for its global coverage and versatility -- it can integrate new indicators and new locations with little effort -- while eliminating collection and format conversion complications. These characteristics are highly advantageous in fast response scenarios.

The most recent global crisis, yet to be resolved, is the COVID-19 coronavirus pandemic. As of April 2021, this outbreak reached an unprecedented scale, with more than three million deaths and more than 140 million confirmed cases\cite{Worldometer}. According to the United Nations' framework for the immediate socio-economic response to COVID-19\cite{UNSDG2020}, this critical period is far more than a health hazard. The socio-economic impact is tremendous and will increase poverty and inequalities globally, jeopardizing lives and livelihoods, especially for vulnerable groups. Other issues include the lack of adequate social protection, losses in income and jobs, increased food insecurity, and a decline in global trades. These problems show the value of acting fast to mitigate adverse effects when a proper response is not timely possible and the recovery extension is unknown. To support such actions, we combine machine learning and satellite data to provide accessible COVID-19 information agilely.

Researchers explored different human signatures visible from space in the literature. Night-time lights disclose urbanization and population levels and can indicate wealthiness\cite{Jean2016} and socio-economic dynamics\cite{Bennett2017}. The tropospheric nitrogen dioxide concentration, primarily affected by fossil fuel consumption, directly correlates with economic activity variations\cite{LiuF2020}. The food supply chain is evaluable through land use classification of agricultural sites\cite{Weiss2020,Alcantara2012} or transportation infrastructure monitoring\cite{Hoppe2016,Tello2006}. Along this line, aircraft are of particular interest to unveil human and economic activities ({\it e.g.}, travel, tourism, freight) and track disease spread due to in-flight transmission\cite{Choi2020,Khanh2020,Pavli2020}.

\begin{figure}[!ht]
\centering
\subfigure[Satellite movement]{
\begin{tikzpicture}[scale=0.6, every node/.style={scale=0.6}]

   \node[inner sep=0pt] (img1) at (0.1,5.6) { \includegraphics[scale=0.25]{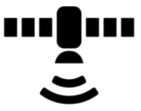}};
   \node[inner sep=0pt] (img2) at (2.25,5.6) { \includegraphics[scale=0.25]{./images/icon_satellite.png}};
   \node[inner sep=0pt] (img3) at (4.4,5.6) { \includegraphics[scale=0.25]{./images/icon_satellite.png}};

   \draw[->, >=stealth, shorten >= 3pt, shorten <= 3pt] (img1) to (img2);
   \draw[->, >=stealth, shorten >= 3pt, shorten <= 3pt] (img2) to (img3);

   \draw[-,color=red,ultra thick] (-0.3,5.0) -- (-0.7,4.5);
   \draw[-,color=red,ultra thick] (0.1,4.9) -- (0.1,4.3);
   \draw[-,color=red,ultra thick] (0.5,5.0) -- (0.9,4.5);
   \draw[-,color=red] (1.04,4.30) -- (1.82,3.3);

   \draw[-,color=green,ultra thick] (1.50,4.5) -- (1.85,5.0);
   \draw[-,color=green,ultra thick] (2.25,4.9) -- (2.25,4.3);
   \draw[-,color=green,ultra thick] (3.05,4.5) -- (2.65,5.0);
   \draw[-,color=green] (2.25,4.20) -- (2.25,3.3);

   \draw[-,color=blue,ultra thick] (3.6,4.5) -- (4.0,5.0);
   \draw[-,color=blue,ultra thick] (4.4,4.3) -- (4.4,4.9);
   \draw[-,color=blue,ultra thick] (4.8,5.0) -- (5.2,4.5);
   \draw[-,color=blue] (2.7,3.3) -- (3.47,4.3);

   \node[inner sep=0pt] (img1) at (2.25,2.9) { \includegraphics[scale=0.15]{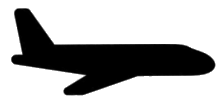}};

   \draw[color=gray!60] (2.25,0.3) ellipse (2.5cm and 0.79cm);

   \node[inner sep=0pt] (img1) at (0.7,0.3) { \includegraphics[scale=0.05]{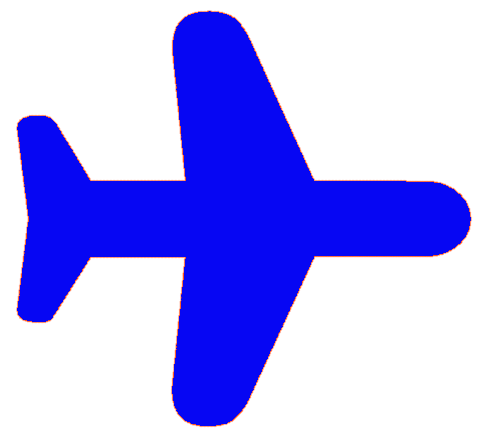}};
   \draw[-,color=blue] (1.82,2.2) -- (1.12,1.3);
 
   \node[inner sep=0pt] (img1) at (2.25,0.3) { \includegraphics[scale=0.05]{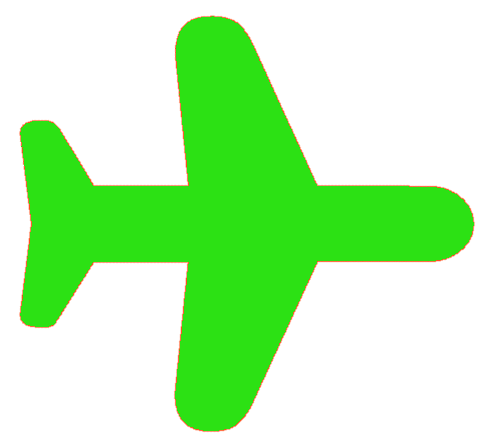}};
   \draw[-,color=green] (2.25,2.2) -- (2.25,1.3);

   \node[inner sep=0pt] (img1) at (3.8,0.3) { \includegraphics[scale=0.05]{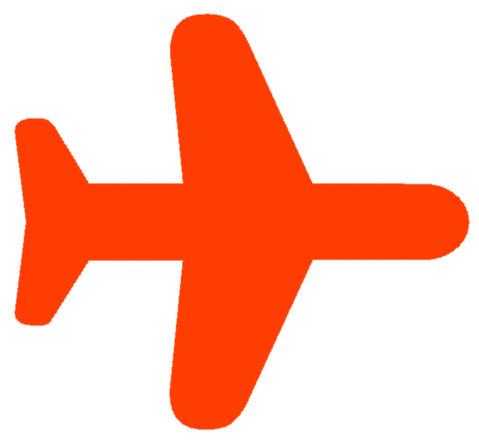}};
   \draw[-,color=red] (2.68,2.2) -- (3.38,1.3);
\end{tikzpicture}
}
\subfigure[Object movement]{
\begin{tikzpicture}[scale=0.6, every node/.style={scale=0.6}]
   \node[inner sep=0pt] (img1) at (2.25,5.6) { \includegraphics[scale=0.25]{./images/icon_satellite.png}};

   \draw[-,color=red,ultra thick] (1.55,4.5) -- (1.85,5.0);
   \draw[-,color=green,ultra thick] (2.25,4.9) -- (2.25,4.3);
   \draw[-,color=blue,ultra thick] (2.95,4.5) -- (2.65,5.0);
   \draw[-,color=red] (1.45,4.3) -- (0.85,3.3);
   \draw[-,color=green] (2.25,4.20) -- (2.25,3.3);
   \draw[-,color=blue] (3.05,4.30) -- (3.65,3.3);

   \node[inner sep=1pt] (plane1) at (0.7,2.9) { \includegraphics[scale=0.15]{./images/icon_plane.png}};
   \node[inner sep=1pt] (plane2) at (2.25,2.9) { \includegraphics[scale=0.15]{./images/icon_plane.png}};
   \node[inner sep=1pt] (plane3) at (3.8,2.9) { \includegraphics[scale=0.15]{./images/icon_plane.png}};

   \draw[color=gray!60] (2.25,0.3) ellipse (3.6cm and 0.79cm);

   \node[inner sep=0pt] (a) at (-0.4,0.3) { \includegraphics[scale=0.05]{./images/icon_plane_r.png}};
   \draw[-,color=red] (0.4,2.3) -- (-0.2,1.3);
 
   \node[inner sep=0pt] (b) at (2.25,0.3) { \includegraphics[scale=0.05]{./images/icon_plane_g.png}};
   \draw[-,color=green] (2.25,2.3) -- (2.25,1.3);

   \node[inner sep=0pt] (c) at (4.9,0.3) { \includegraphics[scale=0.05]{./images/icon_plane_b.png}};
   \draw[-,color=blue] (4.1,2.3) -- (4.7,1.3);
   
   \draw[->, >=stealth] (plane1) to (plane2);
   \draw[->, >=stealth] (plane2) to (plane3);
   
\end{tikzpicture}
}
\subfigure[]{
   \label{fig:parallax-c}
   \begin{tikzpicture}[scale=0.5, every node/.style={scale=0.5}]
   \definecolor{fuchsia}{rgb}{1.0, 0.0, 1.0}
      \tikzset{myptr/.style={decoration={markings,mark=at position 1 with %
    {\arrow[scale=1.5,>=stealth]{>}}},postaction={decorate}}}
      \node[inner sep=0pt] (a) at (5.0,5.0) {\includegraphics[height=5.3cm]{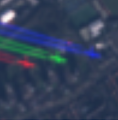}};
      
      \draw[myptr,thick,color=yellow] (3.50,3.90) -- (3.30,2.50);
      \draw[myptr,thick,color=white] (2.70,3.30) -- (4.00,2.90);

      \path[](5.80,3.00) node[yellow,inner sep=1pt]  {\normalsize \textbf{Satellite's heading}};
      
      \path[](5.80,2.60) node[white,inner sep=1pt]  {\normalsize \textbf{Airplane's heading}};
      
   \end{tikzpicture}      
}
\subfigure[]{
   \label{fig:parallax-e}
   \begin{tikzpicture}[scale=0.5, every node/.style={scale=0.5}]
   \definecolor{fuchsia}{rgb}{1.0, 0.0, 1.0}
      \tikzset{myptr/.style={decoration={markings,mark=at position 1 with %
    {\arrow[scale=1.5,>=stealth]{>}}},postaction={decorate}}}
      \node[inner sep=0pt] (a) at (5.0,5.0) {\includegraphics[height=5.3cm]{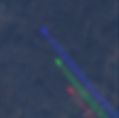}};
      
      \draw[myptr,thick,color=yellow] (3.50,3.90) -- (3.30,2.50);
      \draw[myptr,thick,color=white] (3.80,2.60) -- (3.00,3.90);
      
      \path[](5.80,3.00) node[yellow,inner sep=1pt]  {\normalsize \textbf{Satellite's heading}};
      
      \path[](5.80,2.60) node[white,inner sep=1pt]  {\normalsize \textbf{Airplane's heading}};
      
   \end{tikzpicture}      
}
\subfigure[]{
   \label{fig:parallax-f}
   \begin{tikzpicture}[scale=0.5, every node/.style={scale=0.5}]
   \definecolor{fuchsia}{rgb}{1.0, 0.0, 1.0}
      \tikzset{myptr/.style={decoration={markings,mark=at position 1 with %
    {\arrow[scale=1.5,>=stealth]{>}}},postaction={decorate}}}
      \node[inner sep=0pt] (a) at (5.0,5.0) {\includegraphics[height=5.3cm]{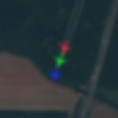}};
      
      \draw[myptr,thick,color=yellow] (3.50,3.90) -- (3.30,2.50);
      \draw[myptr,thick,color=white] (3.60,3.90) -- (3.20,2.50);
      
      \path[](5.80,3.00) node[yellow,inner sep=1pt]  {\normalsize \textbf{Satellite's heading}};
      
      \path[](5.80,2.60) node[white,inner sep=1pt]  {\normalsize \textbf{Airplane's heading}};
      
   \end{tikzpicture}      
}

\caption{Airplane color patterns in Sentinel-2 images: these satellites observe the ground surface at different times in each spectral band, creating (a) a parallax effect for airplanes at high altitudes and (b) a similar color separation effect for airplanes in high-speed at any elevation; both effects combined produce the colored patterns shown in figures (c), (d), and (e). This figure contains modified Sentinel-2 data processed by Euro Data Cube.}
\label{fig:parallax}
\end{figure}

For those reasons, aircraft detection is present in the most advanced aerial scene recognition benchmarks\cite{Lam2018,Xia2018}, and there is extensive literature addressing airport operation, the great majority devoted to stationary aircraft\cite{Zhao2017,KamgarParsi2001,Zhang2016,Grosgeorge2020}. However, the number of parked airplanes is not directly correlatable with airport traffic. For instance, Paris' Charles de Gaule Airport and Rome's Leonardo da Vinci–Fiumicino Airport have, on average, approximately the same number of airplanes on the ground every day\cite{RACEthroughput}, even though the former has about 50\% more flights in the same period\cite{Eurocontrol}. Traffic estimation requires detecting flying aircraft, whose literature is not so developed. Zhao~\emph{et~al.}\cite{Zhao2018} designed ingenious heuristics to perform this task using the water vapor absorption channel from LandSat-8 thanks to the observation that reflectance increases for high-altitude surfaces and generates bright spots on aircraft locations. Despite the high accuracy in ideal conditions, this method is affected by weather conditions ({\it e.g.}, high clouds) and low altitudes ({\it e.g.}, aircraft during take-off and landing). Besides, Landsat-8's repeat cycle of 16 days\cite{USGS2019} hinders the ability to use temporal data to cope with these difficulties. Heiselberg\cite{Heiselberg2019} and Liu~\emph{et~al.}\cite{Liu2020} utilized images captured by the Sentinel-2 satellites for the same job. These satellites' multispectral instrument design makes them observe the earth's surface at different times in each spectral band\cite{SUHET2015}. As the ground serves as a reference to merging bands, the resulting multispectral images (from MSI sensor on-board Sentinel-2) present inter-band measurement displacements due to parallax for objects at high altitudes and high-speed movement for objects at any elevation. Figure~\ref{fig:parallax} illustrates how these displacements create a colored pattern for flying airplanes in the three MSI bands of visible light. The works of Heiselberg\cite{Heiselberg2019} and Liu~\emph{et~al.}\cite{Liu2020} also relied on experts to handcraft heuristics for the airplane detection, which may prevent immediate analyses in outbreaks that require a fast response.

In this work, we take advantage of the advances driven by deep learning algorithms\cite{Lecun2015} -- bio-inspired neural networks that learn representations with multiple abstraction levels and discover intricate patterns in massive data -- to devise a dependable and adaptable detector. Our approach goes beyond the detection of flying airplanes as we use our airplane detector to built a time series of the number of landing or take-off airplanes in airports. We then process those time series to estimate structural breaks (caused by lockdown restriction at the beginning of the pandemic) and to compute the recovery rate for the monitored airports. The presented framework supports decision-making during the COVID-19 pandemic by measuring how fast the airports are recovering toward getting into their normal operation and if such recovery is in accordance with opening and lockdown policies, which should be defined considering the number of cases and deaths of COVID-19 pandemic. Furthermore, our approach was designed to require a small amount of labeled data for training the proposed flying airplane detector, without loss of generality. We validate our approach by monitoring the 30 busiest airports in countries with some integration to the European Union. The results show the effectiveness of our solution to measure such activities and the recovery rate of such airports.

In summary, this study presents several contributions in some subjects relevant to the research field in which this study inserts itself and to the society toward facing the negative aspect of the COVID-19 pandemic. First, this method is currently fully integrated into the \emph{Rapid Action Coronavirus Earth} (RACE) observation dashboard\cite{RACEtraffic}, which is an open platform of the European Space Agency (ESA) that uses Earth observation satellite data and artificial intelligence to measure the impact of the COVID-19 lockdown and to monitor post-lockdown recovery. This study also shows how to design and train a shallow neural network to detect flying airplanes using remote sensing imagery and with a minimum amount of annotated images. Flying airplane detection task is still an open problem in the current literature, and this study contributes with the proposal of a new technique for this task that has been successfully used in a challenging and practical scenario related to combating the COVID-19 pandemic. The third contribution of this study relies on the construction of a new dataset, along with the ground-truth annotations of flying airplanes, to support future researches involving flying airplane detection through satellite images. The fourth contribution refers to the analysis of time series built with our flying airplanes detector to estimate breakouts and recovery rate, automatically, and to support decision-making to opening and closing airports taking into account the official numbers of COVID-19 cases and deaths. Finally, the source code of our solution, the trained models, and the annotated data are freely available to the scientific community, encouraging reproducibility of our results and the use of our solution in similar situations in the future.

\section{Proposed approach}

\tikzfading[name=arrowfading, top color=transparent!0, bottom color=transparent!95]
\tikzset{arrowfill/.style={#1,general shadow={fill=black, shadow yshift=-0.8ex, path fading=arrowfading}}}
\tikzset{arrowstyle/.style n args={3}{draw=#2,arrowfill={#3}, single arrow,minimum height=#1, single arrow,
single arrow head extend=.1cm,}}
\NewDocumentCommand{\tikzfancyarrow}{O{0.8cm} O{black} O{top color=gray!20, bottom color=red!80} m}{
\tikz[baseline=-0.5ex]\node [arrowstyle={#1}{#2}{#3}] {#4};
}

\begin{figure*}[!htb]
\centering
\begin{tikzpicture}[scale=1.0, every node/.style={transform shape}]
  \fill[gray!10!white] (0.0,-1.9) rectangle (15.3,7.5); 
  \node[draw] at (1.75,7.0) {\bf a) AOI selection};
  \draw(1.75,5.0) node[text centered, inner sep=0pt] (a) { \includegraphics[height=3.0cm]{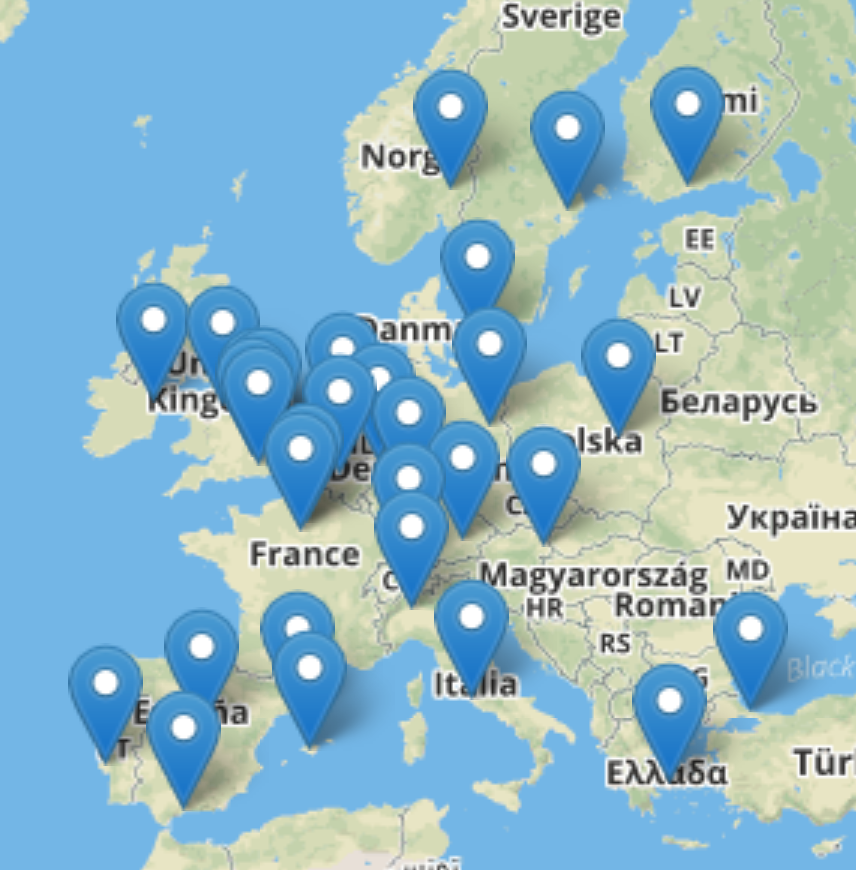}};

  \path[->](3.8,5.0) node[] {\tikzfancyarrow{}};

  \node[draw] at (5.5,7.0) {\bf b) Image acquisition};
  \draw(5.5,5.0) node[text centered, inner sep=0pt] (b) { \includegraphics[height=3.0cm]{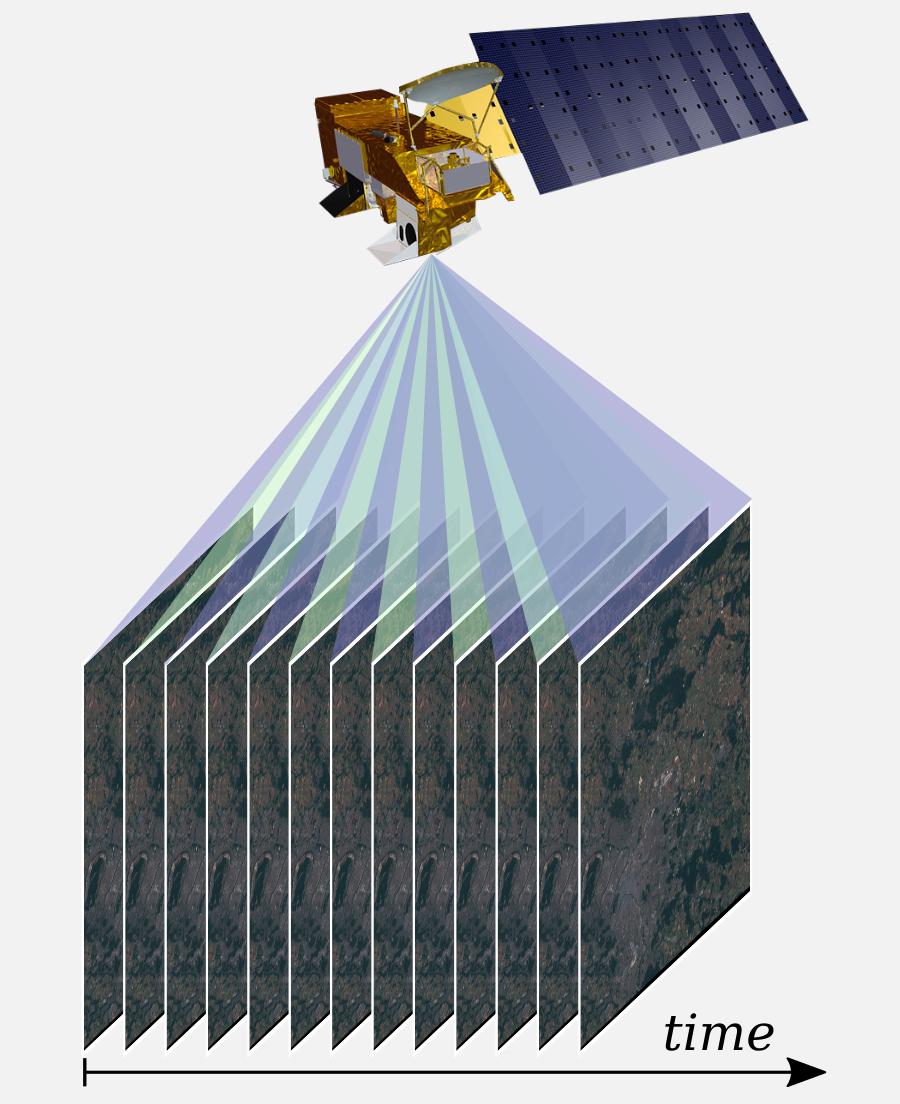}};

  \path[->](6.9,5.0) node[] {\tikzfancyarrow{}};

  \node[draw] at (11.3,7.0) {\bf c) Flying airplane detection};
  \draw(11.3,4.95) node[text centered, inner sep=0pt] (b) { \includegraphics[height=3.0cm]{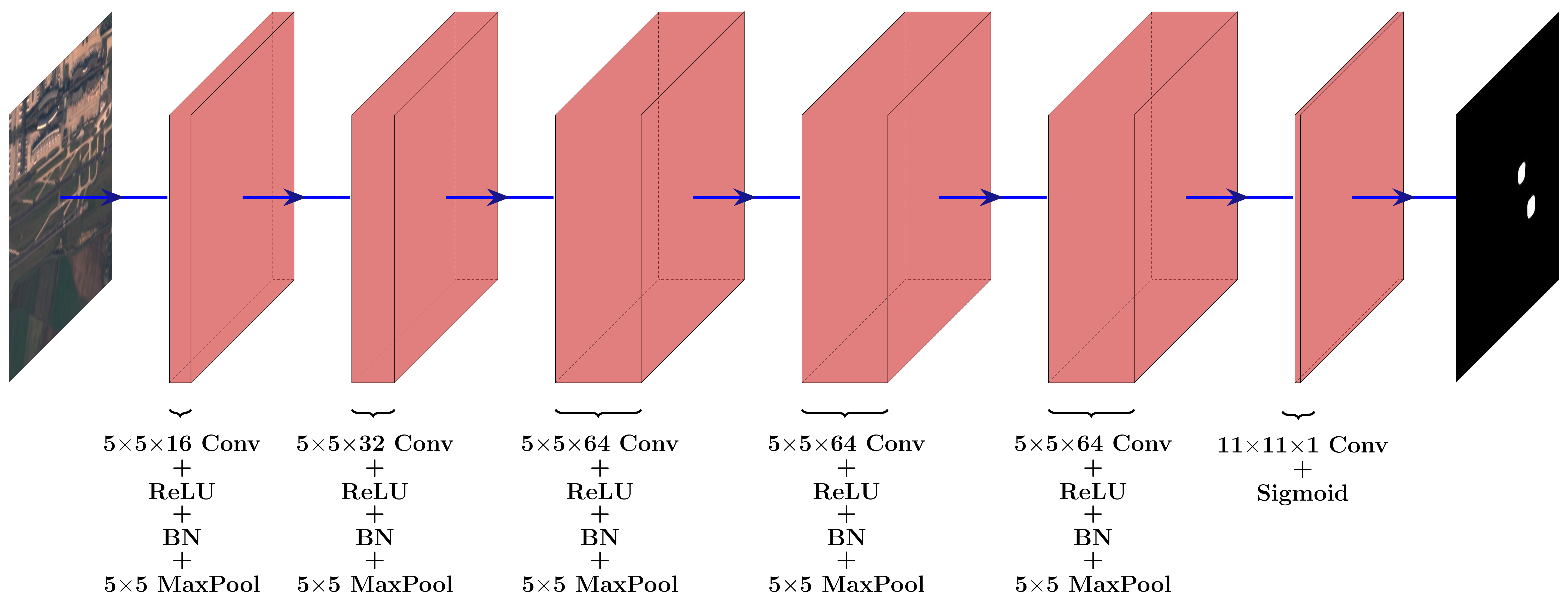}};

  \path[->](12.2,2.8) node[rotate=90,xscale=-1] {\tikzfancyarrow{}};

  \draw(12.2,0.7) node[text centered, inner sep=0pt] (e) { \includegraphics[height=3.0cm]{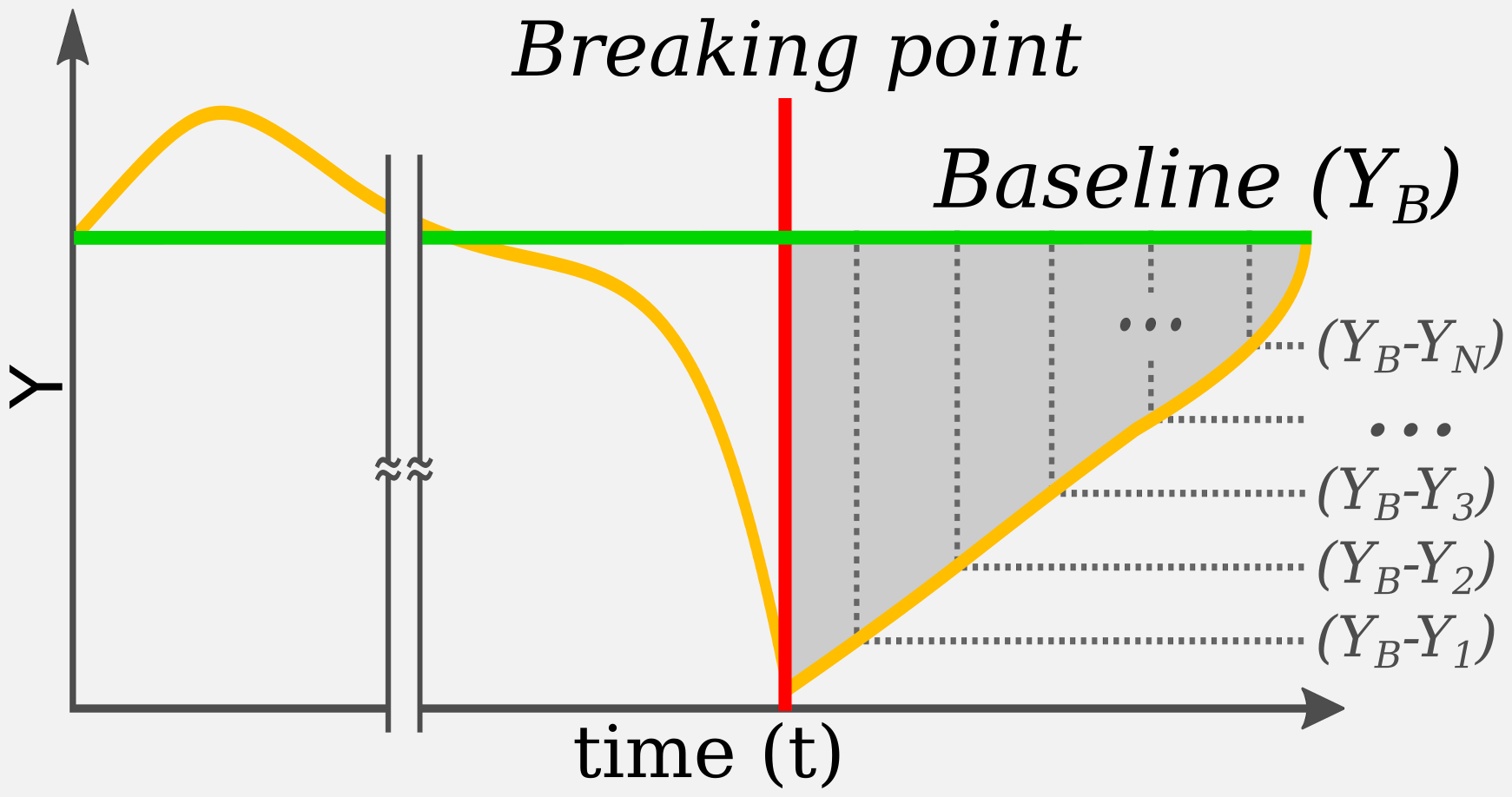}};
  \node[draw] at (12.2,-1.3) {\bf d) Time series analysis};

  \path[->](8.8,0.7) node[xscale=-1] {\tikzfancyarrow{}};

  \draw(6.4,0.7) node[text centered, inner sep=0pt] (f) { \includegraphics[height=3.0cm]{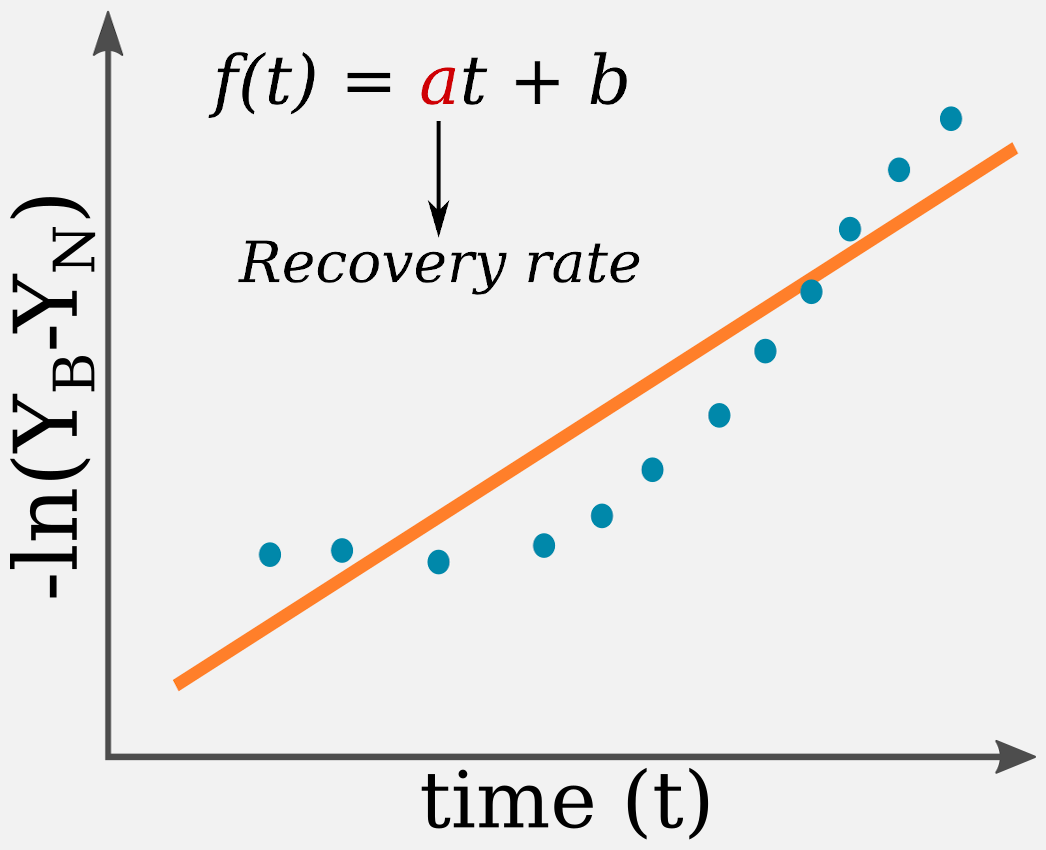}};
  \node[draw] at (6.4,-1.3) {\bf e) Recovery estimation};

  \path[->](4.0,0.7) node[xscale=-1] {\tikzfancyarrow{}};

  \draw(1.80,0.7) node[text centered, inner sep=0pt] (a) { \includegraphics[height=3.0cm]{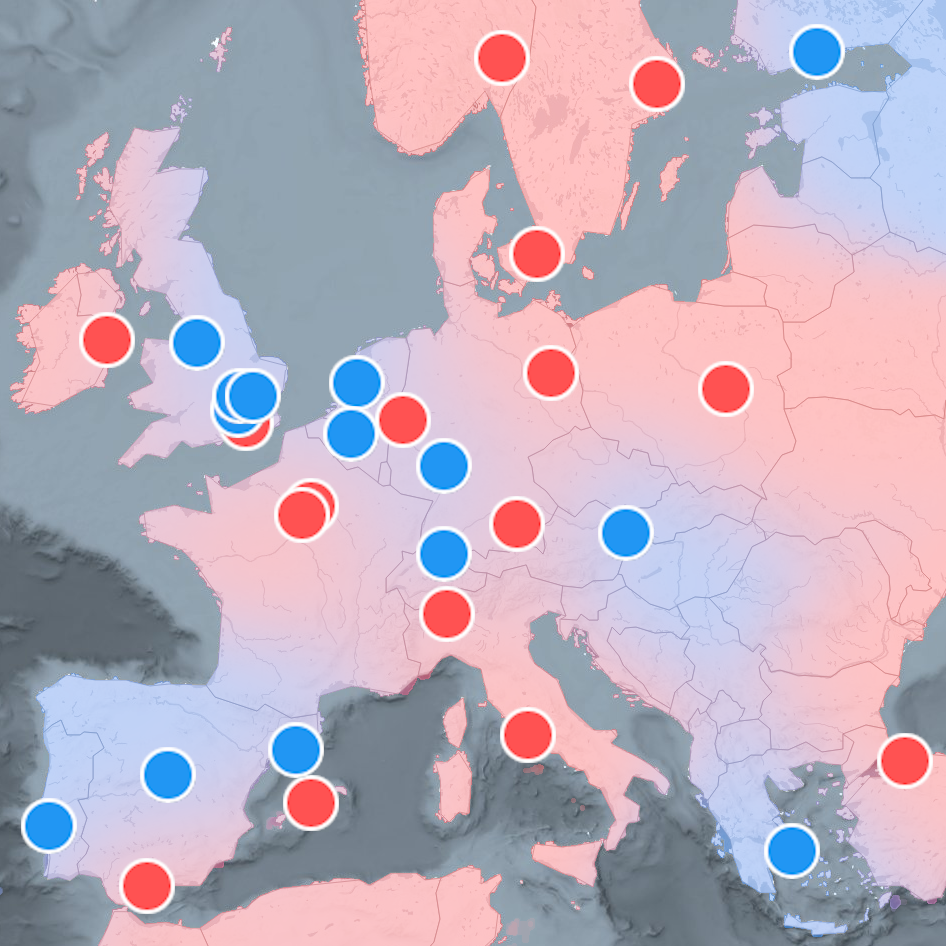}};
  \node[draw] at (1.80,-1.3) {\bf f) Activity indicator};
\end{tikzpicture} 
\caption{The main steps of the proposed approach for detecting flying airplanes and to measure breakouts and recovery rates of airports activities. This figure contains modified Sentinel-2 data processed by Euro Data Cube.}
\label{fig:pipeline}
\end{figure*}
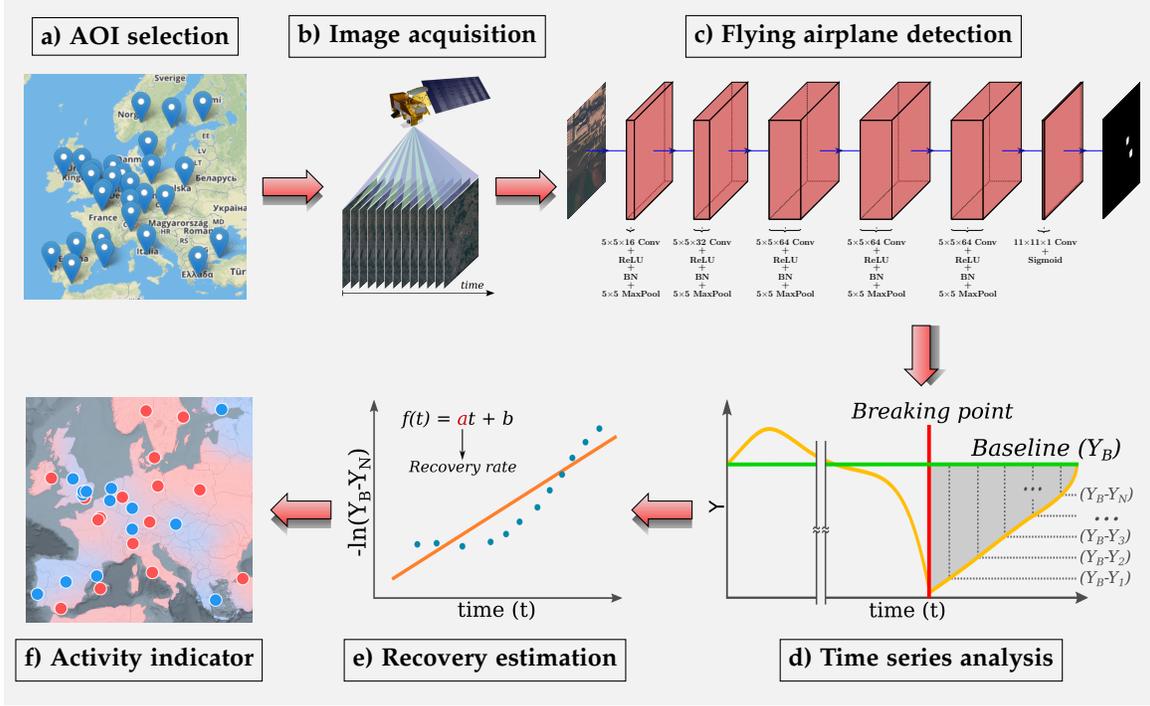

This study introduces a new approach for measuring airport activities using satellite to support decision-making during the COVID-19 outbreak, as illustrated in Figure~\ref{fig:pipeline}. More precisely, we proposed a new method for detecting flying airplanes on Sentinel-2 satellite that enables the monitoring of airport activity in wide areas (Figures~\ref{fig:pipeline}(a)~and~\ref{fig:pipeline}(b)). We adopted the use of shallow fully convolutional networks (FCN) to devise a lightweight architecture (Figure~\ref{fig:pipeline}(c)) containing few trainable parameters and thus requiring lower amounts of training data than other deep learning architectures available in the literature addressing the object detection problem~\cite{Lin2020}. We summarized the detected flying airplanes in a time series representation associated with the number of airplanes around each monitored airport over time (in days). Next, we performed a time series analysis to estimate breaking points associated with the first lockdown restriction related to the COVID-19 pandemic (Figure~\ref{fig:pipeline}(d)). Then, we fitted a log-linear regression model to estimate the recovery rates for the airports considered in this study, which allowed us to correlate the estimated recovery rates with the cases and deaths of coronavirus disease in those locations (Figure~\ref{fig:pipeline}(e)). Finally, we compiled this knowledge into an activity indicator (Figure~\ref{fig:pipeline}(f)) to support the conception, planning, implementation, and evaluation of disease-containment actions. These stages are detailed in the following sections.

\subsection{Data collection}
\label{sec:data_collection}

The first stage of this study consisted of creating a dataset to analyze human travel behavior in the European Union. Our analysis relies on detecting flying airplanes from satellite images and measuring the volume of flights over time. To capture traffic dynamics through remote sensing, the satellite must have a high revisit rate, it must perceive high-speed objects, and imaging resolution must be sufficiently high so that airplanes are visible. The Copernicus Sentinel-2 constellation meets all of these requirements\cite{SUHET2015}. The Sentinel-2 mission includes two identical satellites in the same sun-synchronous polar orbit, $180^\circ$ apart from each other, that revisit any Earth location every 2 to 5 days (higher frequency for areas close to the poles). They capture the visible bands of their multi-spectral images (Red, Green, and Blue (RGB)) with a ground sampling distance (GSD) of 10 meters and a time-lapse of approximately $0.5$ seconds between consecutive bands (red-green and green-blue).

\begin{figure}[!ht]
    \centering
    \subfigure[]{
        \label{fig:datastats-a}
        \includegraphics[width=0.49\textwidth]{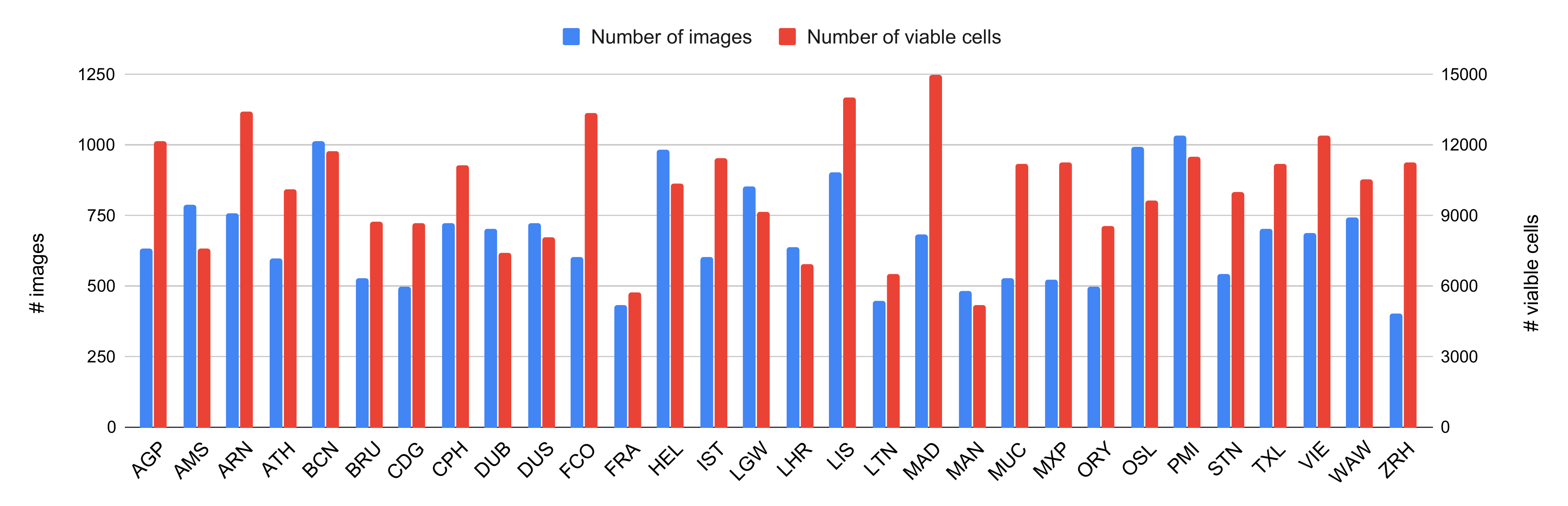}
    }
    \subfigure[]{
        \label{fig:datastats-b}
        \includegraphics[width=0.49\textwidth]{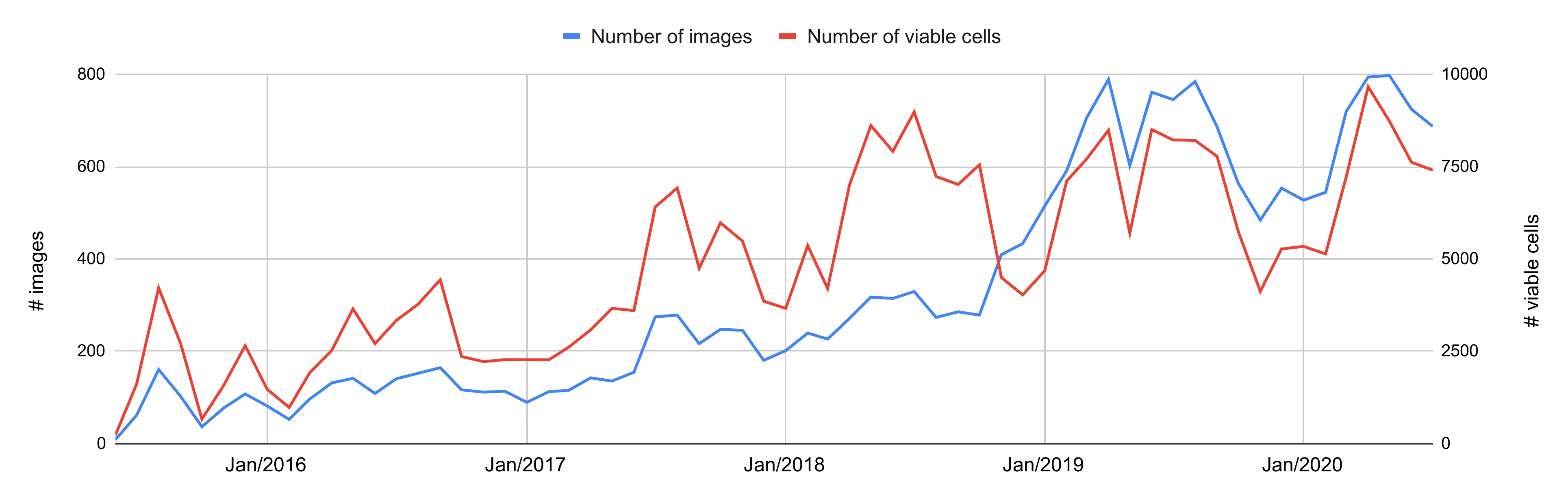}
    }
    \caption{Data collection statistics. The total number of images and viable cells per airport(a) and number of images and viable cells per month for all airports (b).}
    \label{fig:datastats}
\end{figure}

The proposed method operates on sequences of satellite images from Areas of Interest (AOI), which are rectangles with 1.05 longitude degrees in width and 0.7 latitude degrees in height, centered at the geographical coordinates of the 30 busiest airports (i.e., airports that had the highest number of passengers in 2019) in countries with some integration to the European Union. Thus, the AOIs have an area of 6,000 km$^2$, on average, and the selected airports cover 26 different cities and 18 different countries. Figure~\ref{fig:pipeline}(a) highlights their location, and the complete list of International Air Transport Association (IATA) airport codes is shown in Figure~\ref{fig:datastats-a}. The specified AOI provides an observation window of approximately 20 minutes around each airport. More specifically, a satellite image captured at timestamp $t$ shows flight arrivals in the range $[t, t+20min]$ and departures in the range $[t-20min, t]$ for the depicted airport.

We downloaded Sentinel-2 RGB images using the Sentinel-Hub engine\cite{SentinelHub}. We divided each AOI into a $7 \times 7$ grid and evaluated each grid cell's viability considering the following criteria: (1) cells cannot have more than 30\% of cloud coverage; and (2) cells cannot have more than 10\% of missing data. The average number of viable cells per image is 15. Figure~\ref{fig:datastats-a} shows the number of images and viable cells per airport, while Figure~\ref{fig:datastats-b} shows the total number of images and viable cells over time. These images were captured by Sentinel-2 satellites between June 26th, 2015 and July 30th, 2020. 

\subsection{Shallow FCN for detecting flying airplanes}
\label{sec:trainingdetails}

\begin{algorithm*}[t] \small
\caption{Pseudo-algorithm for training our FCN model.}
\label{alg1}
 
\KwIn{Training images, Flying airplane annotations}
\KwResult{FCN model that detects flying airplanes on satellite images}
Randomly initialize the learnable parameters of the FCN model\;
Follow the strategy shown in Figure~\ref{fig:sampling} to build an initial set of training samples by using airplane annotations to extract image patches with $51 \times 51$ pixels of positive and negative samples from the input images; extract additional negative patches from random locations that are at least 25 pixels away from all annotations until a 1:2 ratio between the amount of positive and negative samples is reached\;
Set $score = 0$\;
\Repeat{\normalfont $score$ does not increase for $N$ epochs or the maximum number of epochs $M$ is reached}{
 Run one training epoch (3,000 model update iterations) using: (1)~mini-batches with 256 random samples, (2)~random flips (horizontal and vertical) and 90-degree rotations over samples as data augmentation operations, and (3)~the Adam algorithm with a learning rate of $10^{-4}$ to optimize the binary cross-entropy loss function\;
 Run the current model over the training images and save the list of detected airplanes\;
 Initially mark all detections as false alarms\;
 \ForEach{\normalfont annotated airplane}{
  Find the closest detection to the current annotation and mark this detection as a true positive if the distance between them is less than or equal to 25 pixels (half of the patch size)\;
 }
 Compute the detection rate $DR$ as the number of detections marked as true positives divided by the number of annotated airplanes\;
 Compute the false discovery rate $FDR$ as the number of detections marked as false alarms divided by the total number of detections\;
 Set $score = DR(1-FDR)$, and save the current model if $score$ increases\;
 Update the set of training samples by randomly replacing up to half of the negative patches with $51 \times 51$ patches extracted from detections annotated as false alarms in this iteration\;
}
\end{algorithm*}

Current state-of-the-art detectors either use region proposal or feature pyramid networks to estimate both bounding box coordinates and classes of objects in a scene~\cite{Ren2015, Lin2020}. Such detectors take advantage of deep architectures~\cite{8698456} that contain hundreds of millions of trainable parameters and thus require large training data. With this in mind, we hypotheses the use of shallow architectures is more adequate for modeling our problem due to the absence of a large training data with annotated ground-truth. Shallow architectures have fewer trainable parameters, in comparison to deep architectures, and thus require a reduced amount of labeled data during the training stage. It is important to notice that labeling data is a costly and a time-consuming process, which may become prohibitive when a rapid response is necessary. Furthermore, as we are only interested in counting airplanes, information like bounding boxes and airplane size, in terms of pixels, is not relevant to us. With this in mind, we can reduce the complexity of our architecture by modeling the problem of counting airplanes as a classification problem devised to classify each pixel of the image as being or not the center of a flying airplane (green dot in Figures~\ref{fig:parallax-c}-\ref{fig:parallax-f}).

To validate our hypothesis, we designed a shallow FCN~\cite{Long2015} to produce a probability value for each pixel of an input image, as illustrated in Figure~\ref{fig:pipeline}(c). More precisely, the receptive field around each pixel was set to a region of $51 \times 51$ pixels, which perceives airplanes traveling up to 1,800 kilometers per hour, twice as much as the typical commercial cruise speed. This surplus handles the variability introduced by parallax and altitude changes. The proposed shallow FCN architecture consists of five consecutive $5\times5$ convolutional layers, each followed by a rectified linear unit activation~\cite{Glorot2011}, batch normalization\cite{Ioffe2015}, and a $5\times5$ max pooling. Then, we added an $11\times11$ convolutional layer with sigmoid activation to output values between $0$ and $1$. All layers use unit strides so that the output resolution is the same as the input. This architecture has a total of 277,745 learnable parameters, which correspond to a model size of only $\sim$1.1MB. Finally, non-maximum suppression\cite{Neubeck2006} returns unique detections, and a threshold of $0.5$ selects the ones that most likely represent an airplane. 

Due to the global nature of the COVID-19 pandemic, the generalization is an important aspect to consider. More precisely, it is essential to detect flying airplanes in unknown areas without the need for retraining our model. To evaluate this competence, we used the airports with the first 15 IATA codes in alphabetical order for training (see Figure~\ref{fig:datastats-a}), and remaining 15 airports were used for testing only. With this split we can measure our detector's performance in airports that were not seen during training to validate its generalization capability. Besides, we only used images from January 1, 2020 to June 30, 2020 for training. Thus, we can also use images outside this time period to evaluate our detector's behavior in unseen images from known areas (training airports).

We adopted a semi-automatic strategy to obtain enough annotations even when a rapid response is required. First, we manually annotate as many instances of the target object as possible in a small number of images (in this work, we annotated 190 flying airplanes in 18 images from the Charles de Gaulle Airport). Then, we gradually expand this set of annotations by alternating between training a model with the existing annotations and manually inspecting the detection results to update the annotation set. In our case, we trained an initial model using our 18 annotated images and used it to detect airplanes in all training images. We removed all false alarms from the set of detected airplanes through a visual inspection, and used the coordinates of the remaining detections as the new set of annotations. We repeated this process once more, but this time we also adjusted the detected coordinates to overlay the green dot of the flying airplane pattern (see Figures~\ref{fig:parallax-c}-\ref{fig:parallax-f}). We ended up with 1782 flying airplanes in our final annotation set, which was used to train our final detection model. Every training repetition was carried out by Algorithm~\ref{alg1}, with $N=1$ and $M=\infty$ for temporary models and $N=10$ and $M=50$ for the final detector. $N$ is a patience parameter used to stop the training early if the model does not improve for $N$ consecutive epochs, and $M$ is the maximum number of epochs.

\begin{figure}[!ht]
\centering
\includegraphics[width=0.23\textwidth]{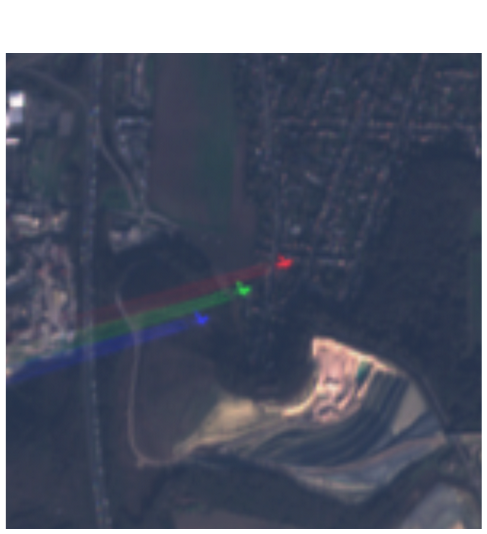}
\includegraphics[width=0.23\textwidth]{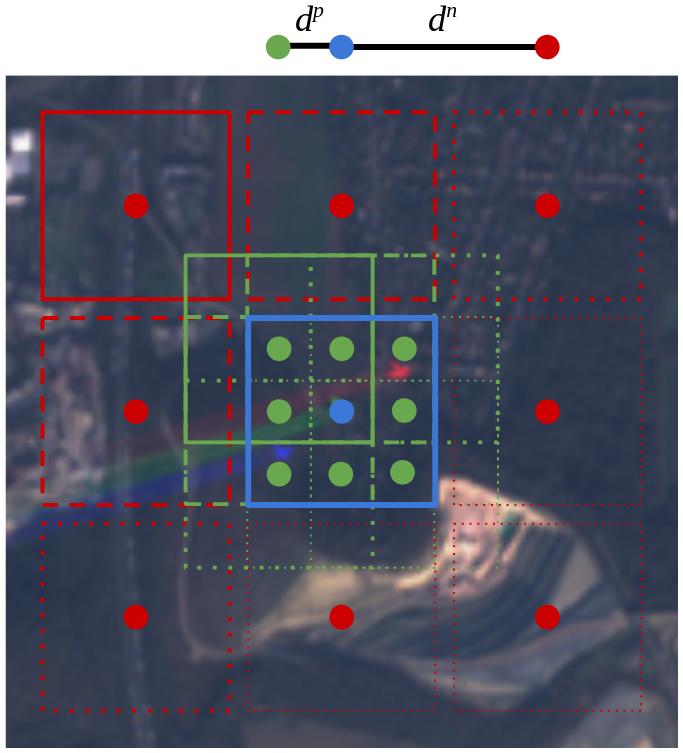}
\caption{Sampling strategy for training: for each annotation (point shown in blue), we extract positive image patches centered on blue and green points. Green points are $d^p$ pixels away from the blue one in one or both axes ($d^p = 3$). Negative image patches are centered on red points, which are $d^n$ pixels away from the blue one in one or both axes ($d^n = 25$). This figure contains modified Sentinel-2 data processed by Euro Data Cube.}
\label{fig:sampling}
\end{figure}

\subsection{Time series generation}
\label{sec:seriesgeneration}

As shown in Figure~\ref{fig:openskythumb}, satellite images do not always cover the entire AOI due to cloud occlusions or to a misalignment between the satellite visible area and the AOI. Thus, estimating the airplane count within a day is not accurate enough for further calculations. We alleviate this problem by using a temporal window $w$. The set of satellite images $I_w$ within this window are used to produce an average airplane count $C_w$, as shown in Equation~(\ref{eq:temporalavg}):
\begin{equation}
    C_w = \sum_{i=1}^7\sum_{j=1}^7\frac{\sum_{k \in I_w} c_{ij}^k}{\max\{1,\sum_{k \in I_w} v_{ij}^k\}}
    \label{eq:temporalavg}
\end{equation}
where $c_{ij}^k$ is the airplane count for the cell in the $i$-th row and $j$-th column of the $k$-th image, and $v_{ij}^k$ is $1$ if this cell is viable and $0$ otherwise. As can be seen, we compute averages at cell level for our $7 \times 7$ AOI grids and then sum all cell averages to obtain a count estimate at image level. 
We use a window size of 30 days with a step size of one day to create our time series (see examples in Figure~\ref{fig:exemple_recovery_rates-c}).

\begin{figure}[!ht]
\centering
\subfigure[]{
    \label{fig:noise-a}
    \includegraphics[width=0.23\textwidth]{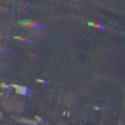}
}
\subfigure[]{
    \label{fig:noise-b}
    \includegraphics[width=0.23\textwidth]{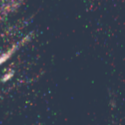}
}
\caption{RGB noise patterns in satellite images caused by (a) misalignment between color bands or by (b) sun-glint. This figure contains modified Sentinel-2 data processed by Euro Data Cube.}
\label{fig:noise}
\end{figure}

In some cases, satellite images may present some artifacts caused by the misalignment between color bands or sun-glint (see Figure~\ref{fig:noise}). Although their occurrence is rare, these artifacts tend to produce several false positives close to each other. If ignored, this problem considerably affects individual cell averages and the final airplane count. To cope with these noisy regions, given that no grid cell in the training set has more than 4 annotated airplanes (considering the final annotation set used for training), we ignore cells that have more than 5 detected airplanes (set $c_{ij}^k=0$ and $v_{ij}^k=0$ if $c_{ij}^k > 5$).

\subsection{Time series analysis}

To estimate the airports' activity in normal and exceptional periods, including the COVID-19 outbreak, we proposed a temporal analysis of time series built from the number of airplanes flying around airports considered in this work. We adopted the use of the concept of the structural breaks, which has been applied in other remote sensing-related problems~\cite{Menini2019GRSM,Roy2019RSE,Bullock2020RSE}. Thus, we characterize the breaks as a position in the time series in which an abrupt shift (or decrease) is observed~\cite{Menini2019GRSM}.

To detect a breaking point in the time series, we adopted two approaches: the simple moving-average (SMA) crossover~\cite{Szakmary2010JBF} and Twitter’s Breakout Detection~\cite{James2016ICBD}. The simple SMA crossover comprises two simple moving-average to follow short-term and long-term tendencies. While the short-term moving average is more reactive to variations, the long-term moving average aggregate changes over a long time and thus produced a smoothed curve. When the curve resultant of these two moving average crosses, then we might have a change of tendency. In turn, the Twitter algorithm employs the E-Divisive with Medians (EDM) method~\cite{Szekely2013JSPI} to automatically detect breakouts in time series. The authors employed the E-statistics to locate changes in mean without any assumption regarding the data distribution. According to authors, this approach was designed to work in presence of anomalies. A more in-depth discussion regarding the mathematics and statistical theories regarding this method can be found in the literature~\cite{James2016ICBD,Szekely2013JSPI}. We evaluate these two algorithms to detect a breaking point in our time series to find the instant $t$ (in days) in which the airport starts recovering from a very low activity state.

After finding the breaking point in time series for each airport, the next stage of our analysis consists of computing the recovery rate of airports' activity~\cite{Nes2007AN}. First, we computed a baseline $Y_{B}$, which is the average of the short-term moving average. Then, we define the recovery rate using the following exponential model~\cite{Veraart2012Nature}, as follows:

\begin{equation}
    \frac{dy}{dt} = -\lambda (Y_{B} - Y_{t})
\end{equation}
\noindent where $Y_{t}$ is the observed trend in the instant $t$ and $\lambda$ is the recovery rate computed by fitting a linear regression of $-ln(Y_{B}-Y_{t})$ against time. 
Figures~\ref{fig:pipeline}(d)~and~\ref{fig:pipeline}(e) illustrate the methodology used to compute the recovery rate. In Figure~\ref{fig:pipeline}(d), for a given time series, the baseline $Y_{B}$ is shown as a green line and the breaking point as a red line. Then, we compute the differences between the baseline and all $Y$ values after the breaking point. These values are transformed into a logarithm scale and used to fit a linear regression (Figure~\ref{fig:pipeline}(e)). Finally, the slope coefficient of the fitted line represents the recovery rate for that time series.

\section{Results}

\subsection{FCN-based detection results}
\label{sec:detectionaccuracy}

To assess the effectiveness of our approach to detect flying airplanes, we designed two experiments in order to measure detection errors in practical scenarios: (1) visual inspections around each detection to determine the proportion of false alarms and (2) comparison to existing publicly available flight records to quantify the occurrence of false negatives. In the context of this work, false alarms occur when our method classifies background patches as a flying airplane, while false negatives occur when the method classifies flying airplane patches as background.

\subsubsection{Visual inspection to quantify false alarms}

We conducted an error analysis to determine the number of false alarms of our method in detecting flying airplanes by visually inspecting the detection results in images that were not seen during training. To this end, we first extracted patches around each detection in images captured between June 2015 and December 2019 from the 15 training AOIs. After manually classifying each patch as a true positive ($TP$) or a false alarm ($FA$), we compute the false discovery rate ($FDR = \frac{FA}{TP+FA}$) for unseen data from AOIs seen during training. We adopted this strategy because it allows accurately estimating FDR in the absence of ground truth annotations. From the total number of 25,747 detections, 410 were false alarms, which corresponds to a $1.59\%$ FDR. This results in $7.6$ false alarms per month, {\it i.e.}, less than one false alarm per airport each month. Knowing that each AOI has an average of 10 images per month and each image has up to 20 million airplane candidates, the incidence of false alarms in areas observed during training is minimal.

To evaluate the generalization capability of our detector operating in unknown areas, we repeated the previous analysis using the images captured between June 2015 and June 2020 from the AOIs not seen during training. There were 616 false alarms in the midst of 30,958 detections, a $1.99\%$ FDR. The average number of false alarms per month was $10.1$, which again resulted in less than one false alarm per airport each month. These values are similar to the ones reported for training AOIs and show that our detector is equally applicable to AOIs that were not seen during training but have a similar setup.

\begin{figure}[!ht]
\centering

\setlength{\tabcolsep}{1pt}
\begin{tabular}{m{0.6cm}m{1.6cm}m{1.6cm}m{0.6cm}m{1.6cm}m{1.6cm}}
(a) &
\includegraphics[width=1.6cm]{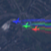} &
\includegraphics[width=1.6cm]{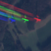} &
(b) &
\includegraphics[width=1.6cm]{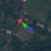} &
\includegraphics[width=1.6cm]{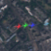} \\
(c) &
\includegraphics[width=1.6cm]{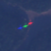} &
\includegraphics[width=1.6cm]{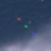} &
(d) &
\includegraphics[width=1.6cm]{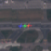} &
\includegraphics[width=1.6cm]{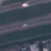}  \\
(e) &
\includegraphics[width=1.6cm]{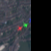} &
\includegraphics[width=1.6cm]{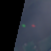} &
(f) &
\includegraphics[width=1.6cm]{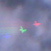} &
\includegraphics[width=1.6cm]{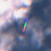}  \\
(g) &
\multicolumn{2}{m{3.3cm}}{\includegraphics[width=3.3cm]{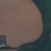}} &
(h) &
\multicolumn{2}{m{3.3cm}}{\includegraphics[width=3.3cm]{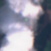}} \\
(i) &
\includegraphics[width=1.6cm]{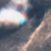} &
\includegraphics[width=1.6cm]{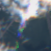} &
(j) &
\includegraphics[width=1.6cm]{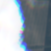} &
\includegraphics[width=1.6cm]{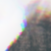} \\
(k) &
\includegraphics[width=1.6cm]{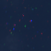} &
\includegraphics[width=1.6cm]{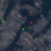} &
(l) &
\includegraphics[width=1.6cm]{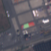} &
\includegraphics[width=1.6cm]{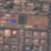} \\
(m) &
\includegraphics[width=1.6cm]{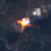} &
\includegraphics[width=1.6cm]{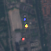} &
(n) &
\includegraphics[width=1.6cm]{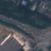} &
\includegraphics[width=1.6cm]{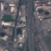} \\
\end{tabular}

\caption{Examples of satellite image patches surrounding automatic detections. Samples classified as \textbf{true positives}: (a-b) large airplanes, whose shape contours are visible, some of them with contrails; (c) medium airplanes, which are still highly visible but have no shape cues; (d) airplanes at low-speed (captured during landing or take off); (e) incomplete airplanes due to missing data; (f) moderate cloud occlusions; (g) small airplanes, which are barely visible even after further magnification; (h) severe cloud occlusions. Samples classified as \textbf{false alarms} caused by: (i) image stitching in cloud areas; (j) parallax in cloud edges;  (k) sun-glint over water surface; (l) colored buildings; (m) ground lights; and (n) movement patterns from vehicles in roadways. This figure contains modified Sentinel-2 data processed by Euro Data Cube.}
\label{fig:det}
\end{figure}

We show several examples of regions depicting correct detections and false alarms in Figure~\ref{fig:det}. These examples show the robustness of our detector to a wide range of variations and illustrate the most common causes of misdetection.

\subsubsection{Analysis of publicly available of flight records to quantify false negatives}

Given the vast amount of data collected in this work, a comprehensive annotation of flying airplanes that allows a precise estimation of the number of false negatives is unfeasible. Thus, to quantitatively measure the detection accuracy, we used publicly available flight track records from the OpenSky Network\cite{OpenSky} as a reference. To do so, we retrieved all records whose portrayed airplanes are inside one AOI at the same moment that the area is imaged by a Sentinel-2 satellite from January 2020 to June 2020 (older records were not available at OpenSky).

\begin{figure}[!ht]
\centering
\subfigure[]{
    \label{fig:openskythumb-a}
    \includegraphics[width=0.23\textwidth]{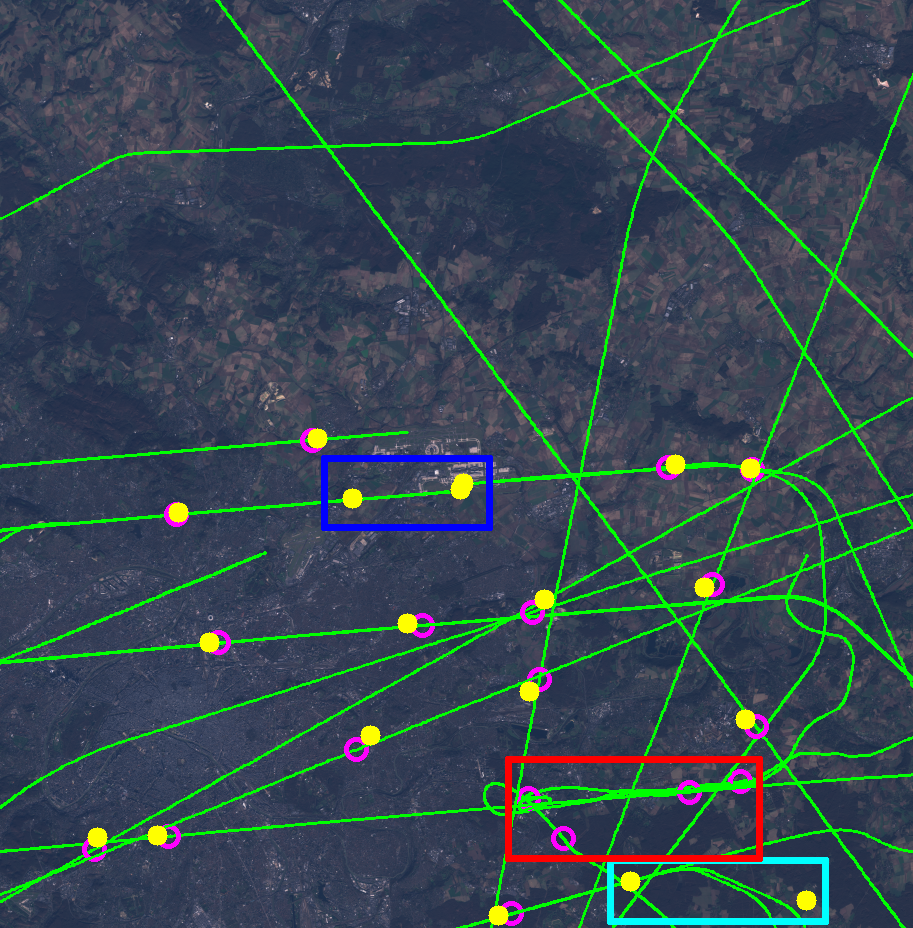}
}
\subfigure[]{
    \label{fig:openskythumb-b}
    \includegraphics[width=0.23\textwidth]{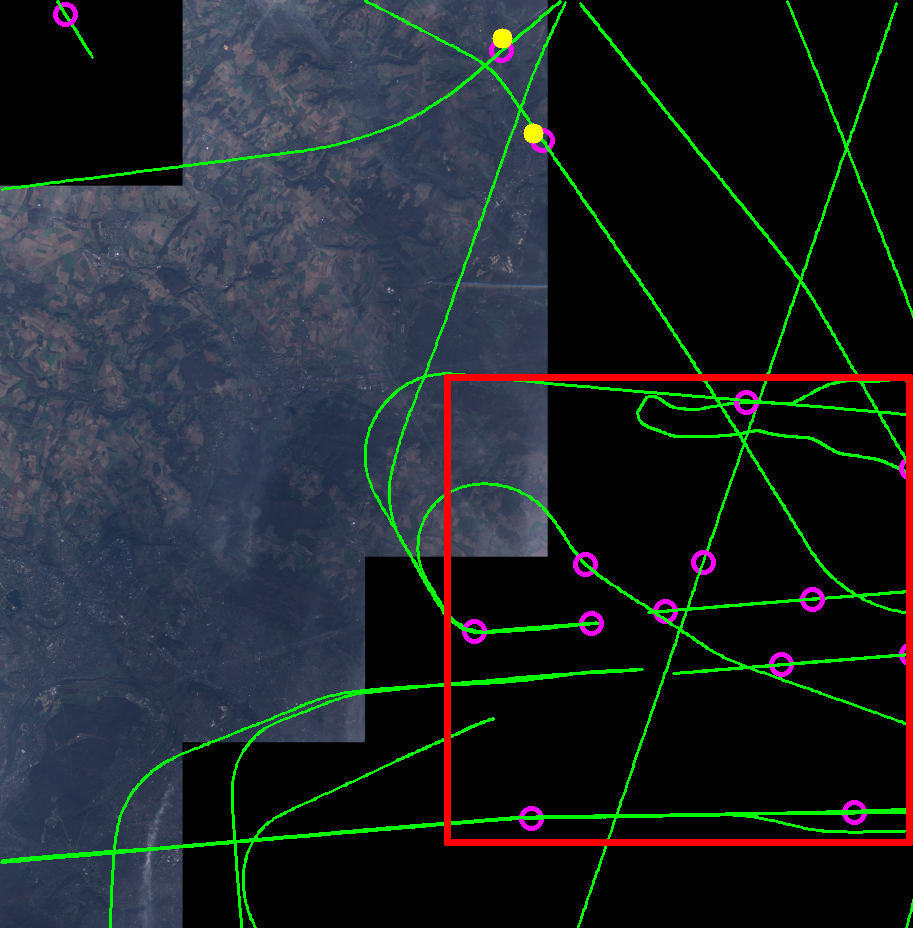}
}
\caption{Visual comparison between our detection results (filled yellow circles) and the OpenSky Network flight records (unfilled pink circles). The green lines represent the OpenSky tracks, from which we trace the location of the airplanes at the image acquisition timestamp. Many airplanes are depicted in both circle styles, and the displacement between yellow and pink circles is caused by different factors ({\it e.g.}, airplane altitude and speed, satellite viewpoint). Sometimes detected airplanes do not have a corresponding OpenSky location due to incomplete tracks during take-off and landing (blue rectangle in (a)) or to missing track records (cyan rectangle in (a)). In turn, some OpenSky locations were not detected by our method because airplanes were too small (red rectangle in (a)) or because they were located over non-viable cells (red rectangle in (b)). This figure contains modified Sentinel-2 data processed by Euro Data Cube.}
\label{fig:openskythumb}
\end{figure}

As shown in Figure~\ref{fig:openskythumb}, these records not necessarily include all airplanes that appear in one image, either due to the absence of tracks for some flights or to the registering of incomplete tracks. Also, not all airplanes whose tracks are available within a certain AOI can be detected by our approach. In most cases, this occurs because these airplanes appear over non-viable cells. In other cases, small airplanes are not visible in the satellite image.

\begin{figure}[!ht]
\centering
\subfigure[]{
 \label{fig:opensky-a}
 \includegraphics[width=0.23\textwidth]{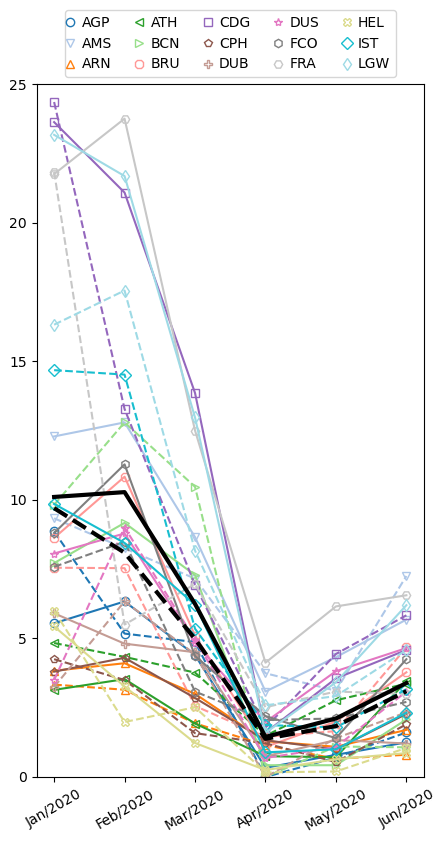}
}
\subfigure[]{
 \label{fig:opensky-b}
 \includegraphics[width=0.23\textwidth]{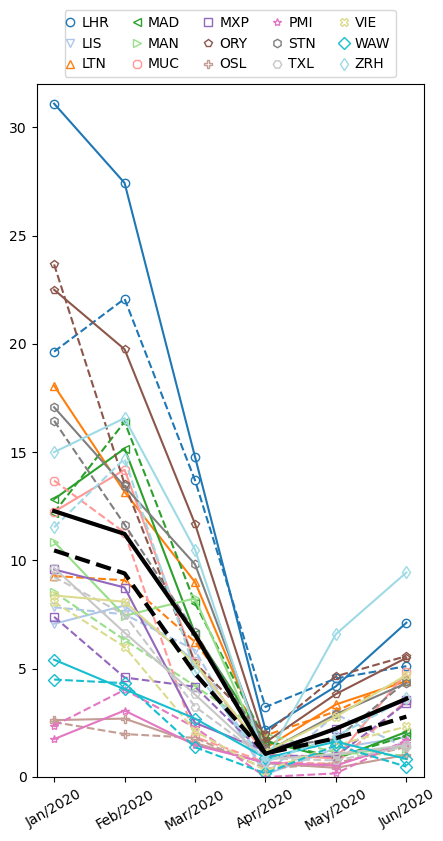}
}
\subfigure[]{
    \label{fig:yoy-a}
    \includegraphics[width=0.49\textwidth]{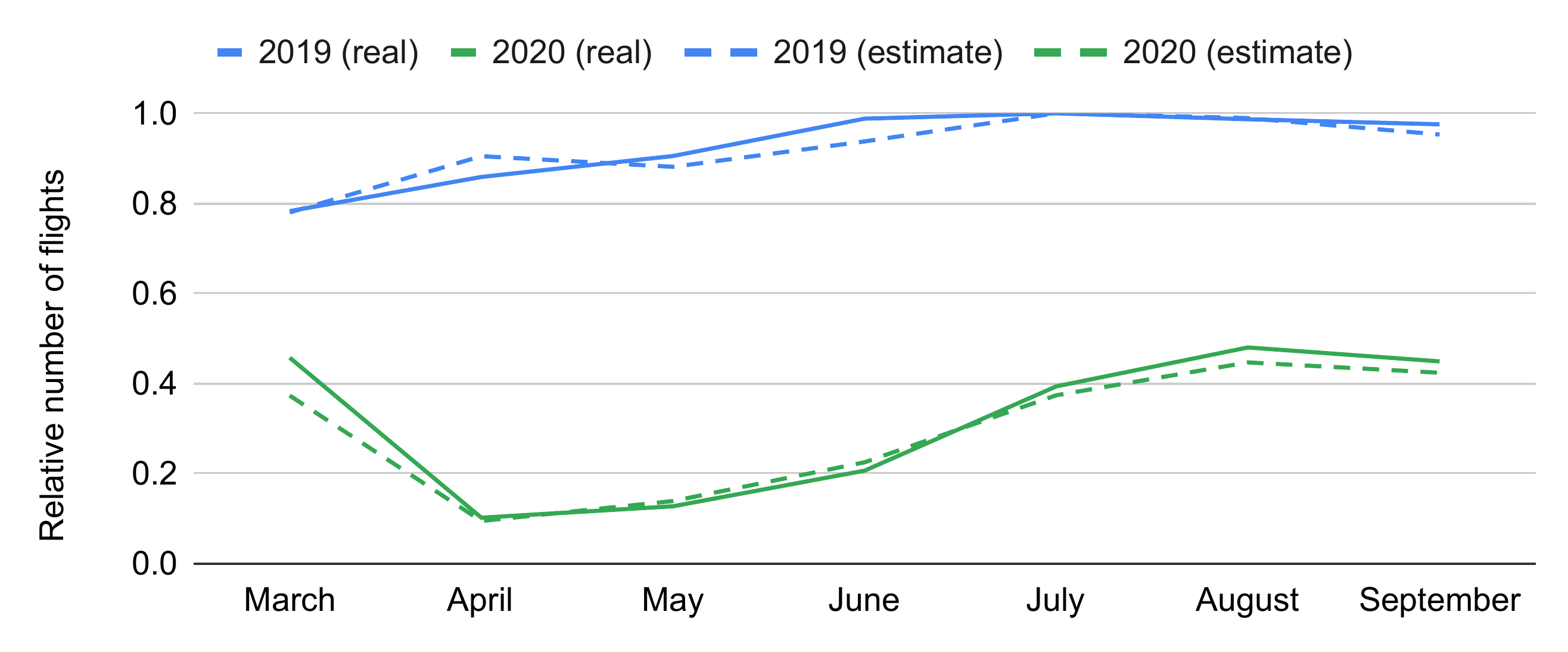}
}\\
\caption{Average number of flights per month computed using our detector (dashed lines) and OpenSky records (solid lines) for (a) known and (b) unknown airports; thick black lines show the average of all airports in the same chart. (c) Relative number of flights from March to September of 2019 and 2020 computed using Eurocontrol's data for 41 EU countries (solid lines) and our framework for 30 airports (dashed lines).}
\label{fig:opensky}
\end{figure}

For this analysis, we created two time series with monthly estimates of the number of flights for each AOI, one using our detection results and the other using the OpenSky records. The time series for training and testing AOIs are presented in Figures~\ref{fig:opensky-a}~and~\ref{fig:opensky-b}, respectively. When comparing the two series of the same AOI, we obtained a Root Mean Squared Error (RMSE) of $2.9$ and a Mean Signed Deviation (MSD) of $-0.8$ on average for training AOIs, and a $2.7$ RMSE and a $-1.4$ MSD on average for testing AOIs. Once again, the results for known and unknown AOIs are close and reinforce our detector's generalization power. The RMSE values show that, when looking into individual AOIs, our results are comparable to OpenSky records, exhibiting similar volume of flights and trends in airport activity. The MSD values reveal that our estimates are consistently lower than their respective OpenSky values. This negative bias approximates the average number of false negatives per image. These false negatives, however, can either be caused by detection failure (Figure~\ref{fig:openskythumb-a}) or data absence (Figure~\ref{fig:openskythumb-b}).

If we consider the average time series from several AOIs (thick black lines in Figures~\ref{fig:opensky-a}~and~\ref{fig:opensky-b}), the RMSE is reduced by two thirds for known airports (from $2.9$ to $1.0$) and by half for unknown airports (from $2.7$ to $1.4$). By averaging multiple AOIs from different countries, we not only obtain stabler series but also integrate airport activity at continental level. To illustrate how accurate those measurements can be, we recreate the real number of flights per month from 42 European countries (data from March to September of 2019 and 2020 made available by Eurocontrol\cite{Eurocontrol}) using the monthly average of the 30 AOIs considered in this work (data from August to September of 2020 generated by the live version of our approach at the RACE dashboard\cite{RACEtraffic}). As both series are in different scales, we normalized them by their maximum values in the considered period, and the outcome is presented in Figure~\ref{fig:yoy-a}. This result shows that our monthly estimates are directly proportional to the real number of flights. Thus, despite noisy, our time series for individual AOIs fluctuate around their expected values and allow us to conduct AOI-specific analyses.

\subsection{Time series analysis results}

This section presents the time series analysis applied in this study to discover outbreaks in airports' activities caused by the COVID-19 pandemic and to estimate their recovery rate. In both analyses, we use the proposed flying airplane detection algorithm to build a time series associated with the number of flying airplanes in each AOI considered in this study.

\subsubsection{Structural break detection}

We assess the effectiveness of both SMA crossover~\cite{Szakmary2010JBF} and Twitter's Breakout Detection~\cite{James2016ICBD} by analyzing their parameter space and then selected the best technique for our problem considering their best configuration. To measure the effectiveness of such methods to detect structural breaks, we considered the date May 1st as being the observed breaking point, which reflects the period in which the airports presented a significantly reduction in their operations. We measure the ability of both techniques in detecting these breaking points by computing the Mean Absolute Error (MAE) and the RMSE between the observed and predicted breaking points.

For the SMA crossover method, we analyzed the short-term parameters considering values ranging from 7 to 49 days and the long-term parameters taking values ranging from 14 to 98 days. From these experiments, we could observe that SMA crossover method presented a better performance using a 14-day window size for the short-term parameter and a 49-day window size for the long-term. For the Twitter algorithm, we analyzed its two main parameters, \textit{msize} and \textit{beta} parameters. For the \textit{msize}, we consider eight values ranging from 64 to 128 days, and for the \textit{beta} parameter (penalization parameter), we consider ten values ranging from $0.1$ to $1.0$. Before applying the twitter algorithm for detecting breaking points, we first use a smoothing technique to remove random variations and thus reveals underlying trends clearly. From these experiments, we could observe a better performance considering a \textit{msize} and \textit{beta} parameter values of 64 and 0.2, respectively. As a result, we observed a $19.7$ MAE and a $23.4$ RMSE for the SMA crossover method, and a $19.9$ MAE and a $26.5$ RMSE for the Twitter algorithm. Both metrics indicate the prediction quality, with errors ranging from 0 to infinity and lower values being better.

Once the algorithms returned quantitative values, we adopted the use of the Wilcoxon Signed-rank (WSR) test statistic to verify if both algorithms are statistically different. The WSR test is a non-parametric test used to assess the null hypothesis that two related paired samples come from the same distribution~\cite{Devore1987Brooksy}. More precisely, we computed the breaking points using both algorithms for the 30 airports considered in this study. Then, we converted the detected breaking point dates into the Julian format, and then we applied the WSR test to check if both algorithms are statistically different. The obtained p-value confirmed that the differences between the two algorithms' results are statistically significant, considering a confidence level of 95\%. Thus, from hereon we use the best configuration of the SMA crossover since this method presented a better performance.

\subsubsection{Recovery rate and COVID-19 analysis}
\label{sec:recover-rate}

To compute the recovery rate, we fitted a log-linear regression model considering a linear regression algorithm with a mean squared error (MSE) as a cost function. In short, we estimated the structural break for each time series and their respective baseline, which corresponds to the average number of flying airplanes. Then, we computed the log of the difference between the daily estimations of counted airplanes and the baseline value. The estimation of log-linear regression models considers such differences (in log scale) as a dependent variable and the data timestamp as an independent variable.

To measure the goodness of fit of regression models to estimate the recovery rate for the airports, we adopted the R-squared metric, also known as coefficient of determination, which ranges from 0.0 to 1.0 and measures the proportion of variance in the independent variable explained by dependent variables (Equation~\ref{eq:r_squared}):
\begin{equation}
    \textit{R-squared} = 1 - \frac{\sum_{t=1}^{n} (y_{t} - \hat{y}_{t})^2}{\sum_{t=1}^{n} (y_{t} - \bar{y})^2}, \qquad \bar{y} = \frac{1}{n} \sum_{t=1}^{n} y_{t}
    \label{eq:r_squared}
\end{equation}
\noindent where $y_t$ and $\hat{y}_t$ are the observed and predicted values for the time $t$. In this context, an R-squared of 0.0 means that the fitted model does not explain any variation in the independent variable around its mean, while an R-squared of 1.0 means that the fitted model explains all variations in the independent variable around its mean, \emph{i.e.}, the obtained regression model fitted all data points~\cite{Devore1987Brooksy}.

\begin{figure}[!ht]
    \centering
    \subfigure[]{
        \includegraphics[width=0.49\textwidth]{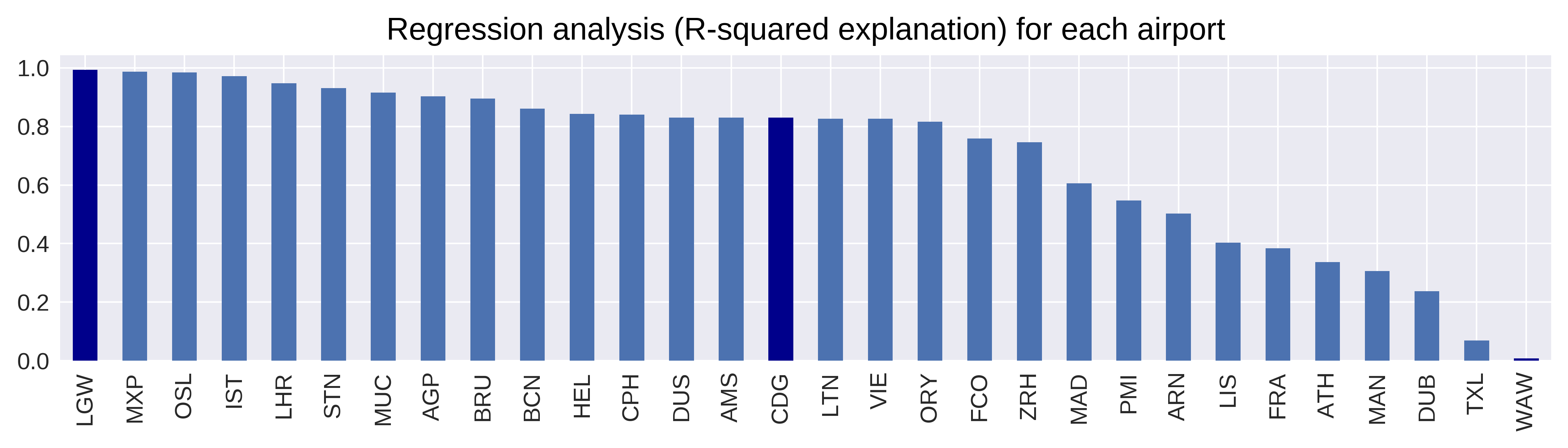}
    }
    \subfigure[]{
        \includegraphics[width=0.16\textwidth]{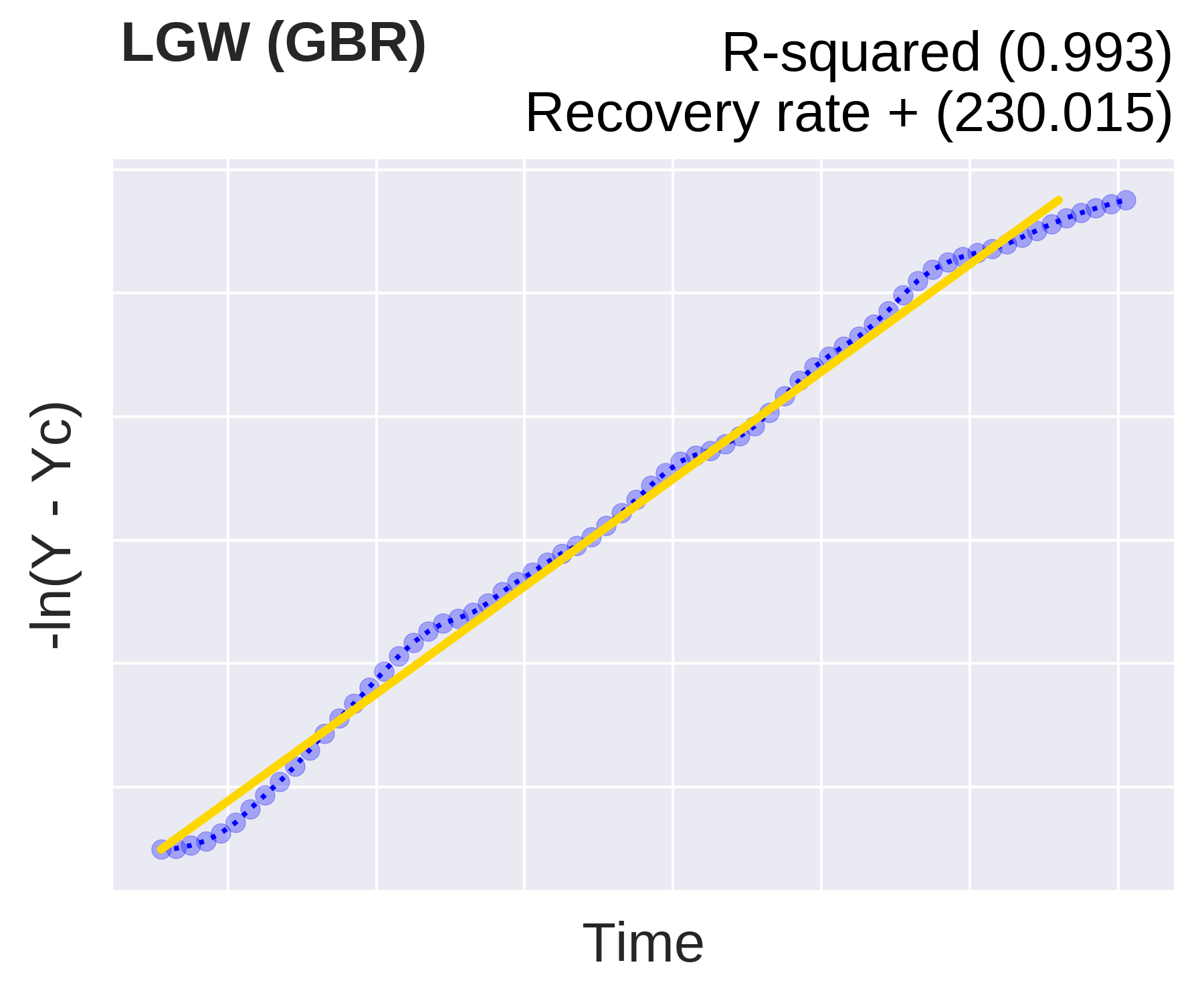}
        \includegraphics[width=0.16\textwidth]{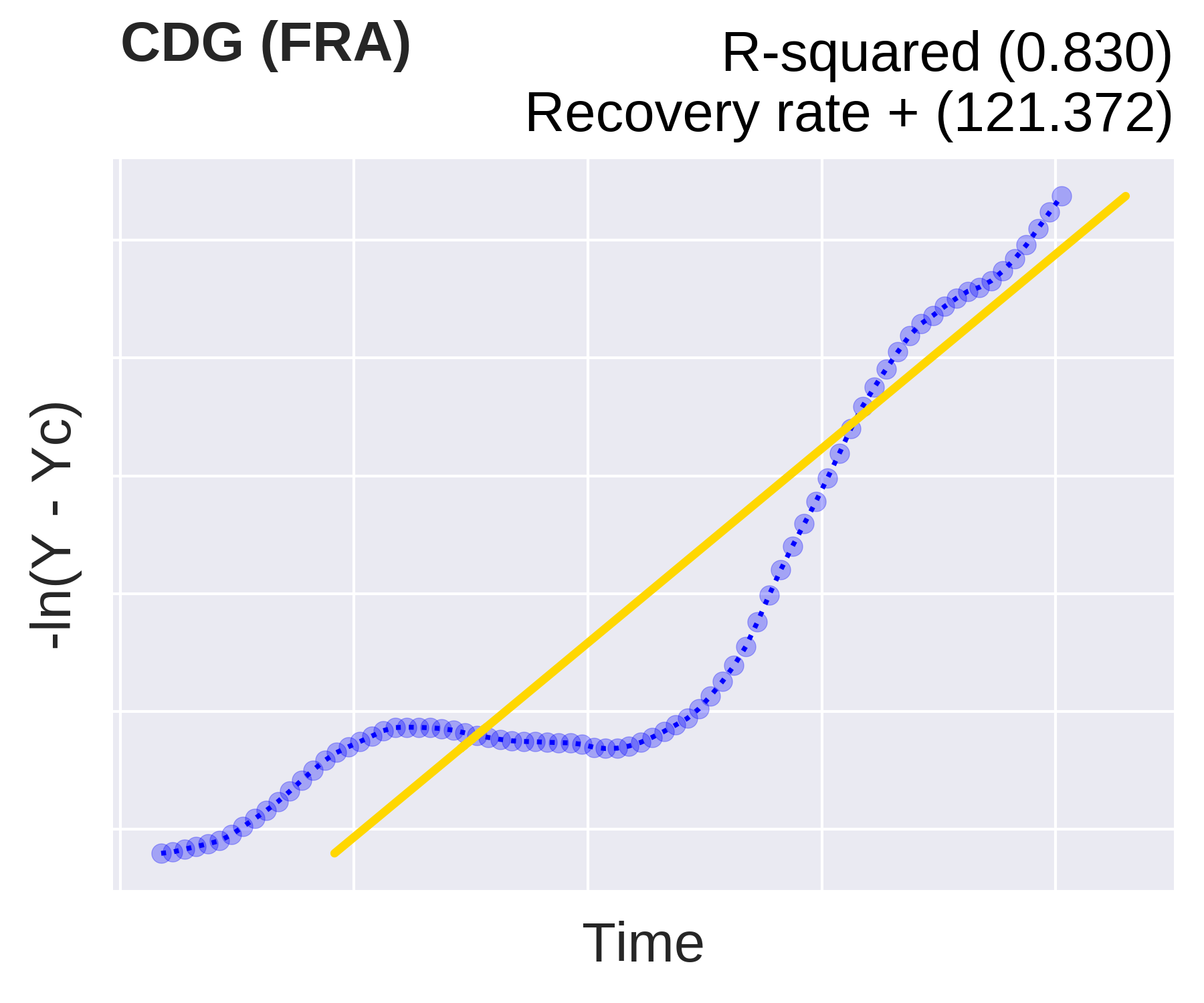}
        \includegraphics[width=0.16\textwidth]{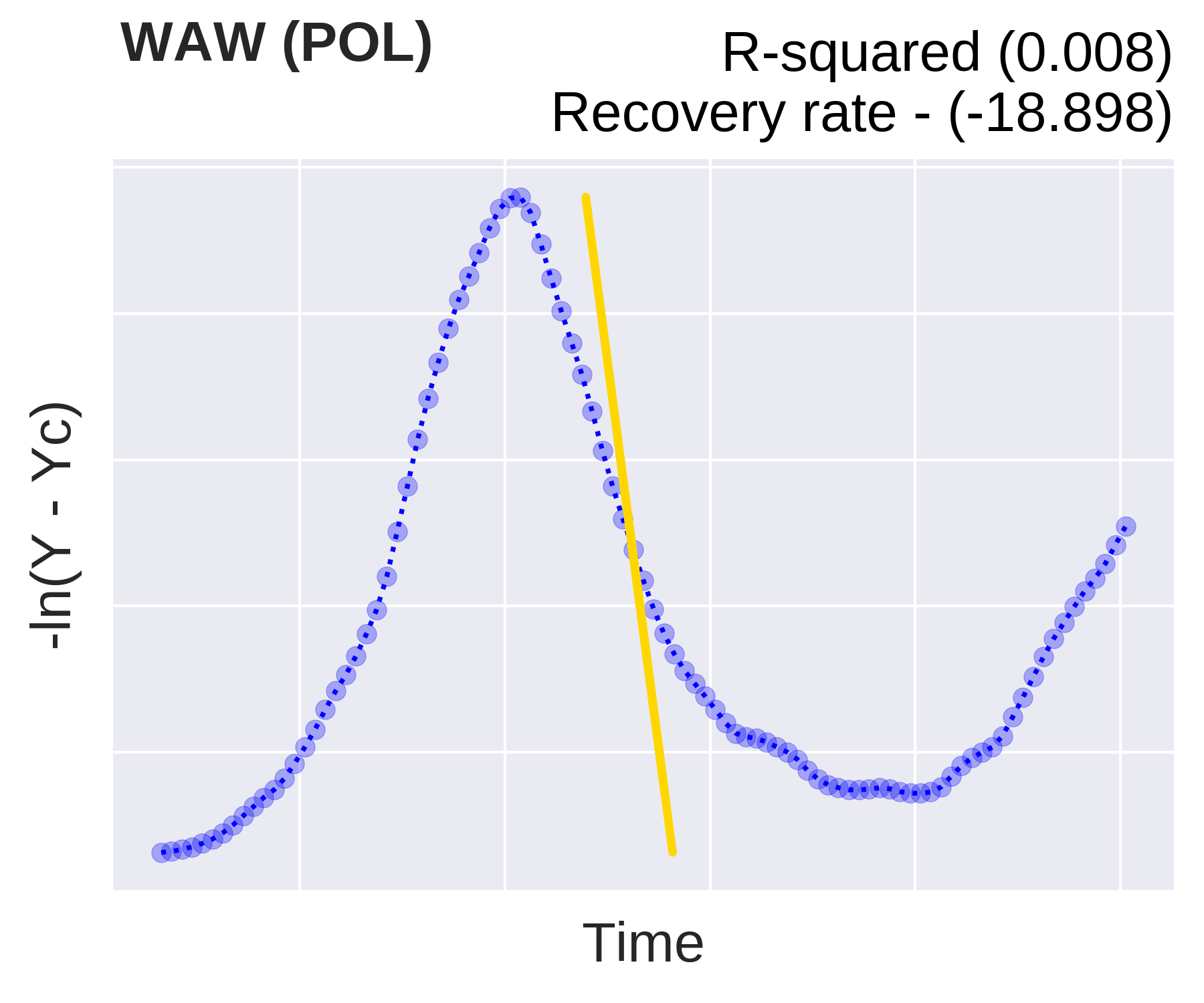}
    }
    \subfigure[]{\label{fig:exemple_recovery_rates-c}
        \includegraphics[width=0.16\textwidth]{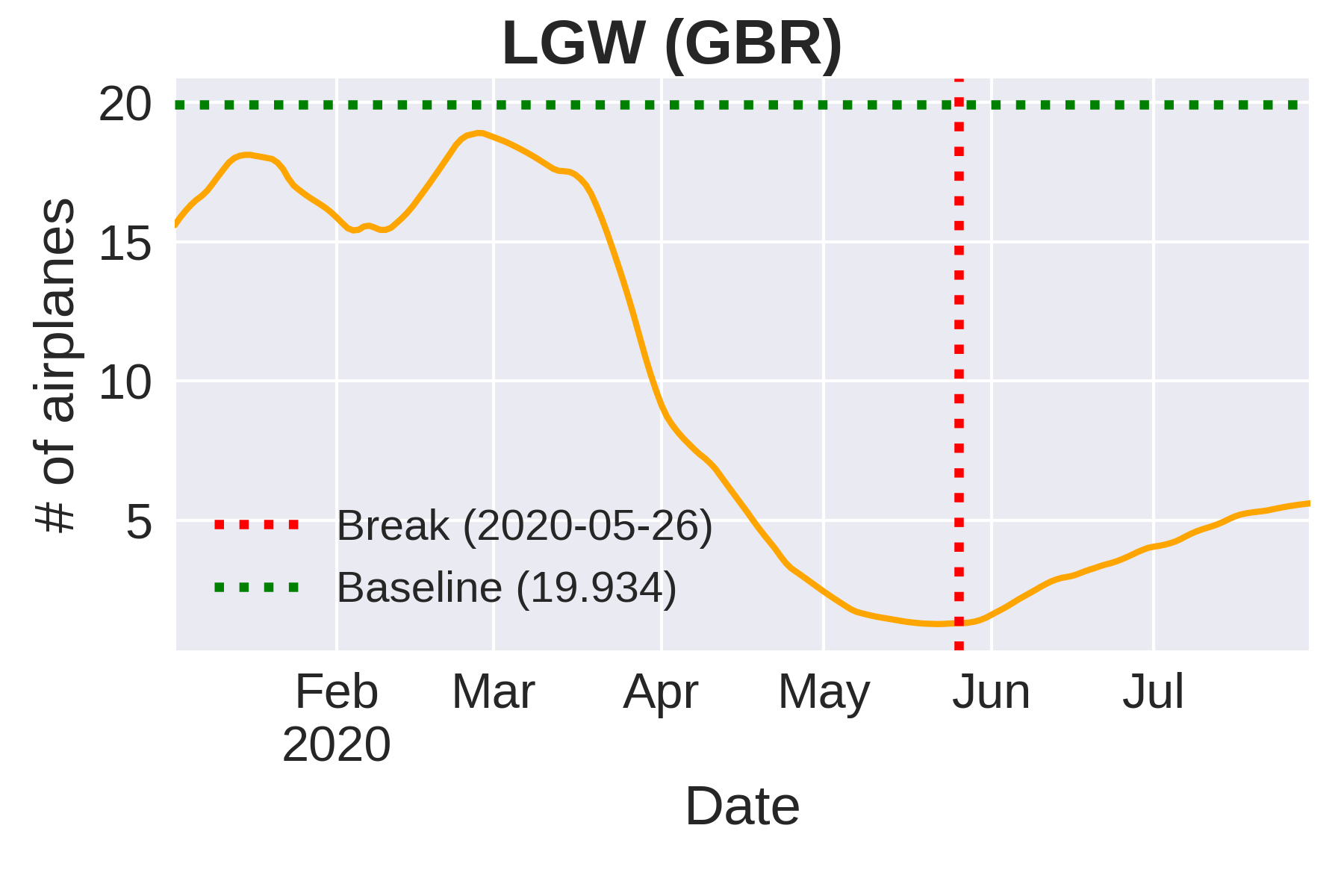}
        \includegraphics[width=0.16\textwidth]{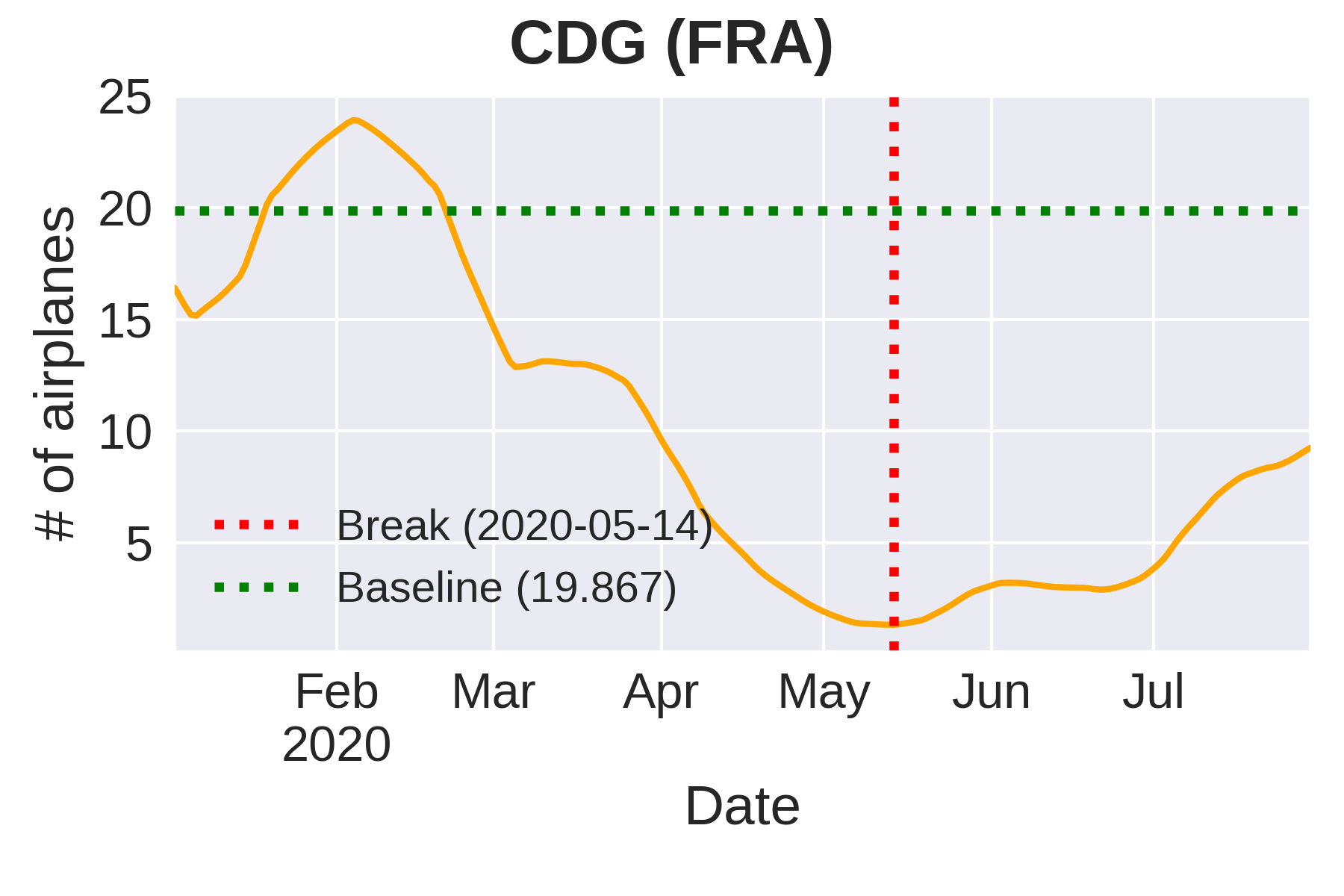}
        \includegraphics[width=0.16\textwidth]{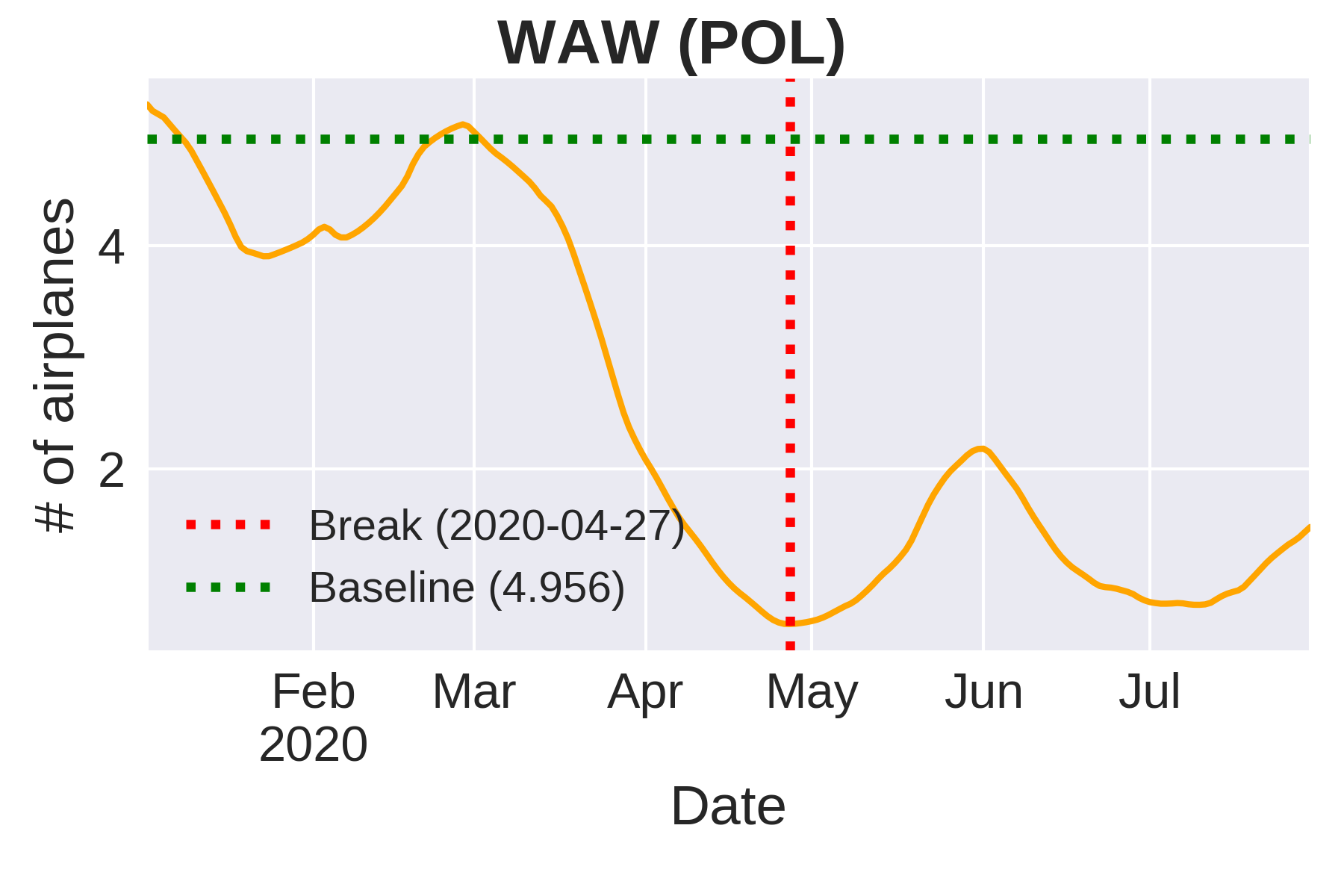}
    }
    \caption{The top figure shows (a) the performance results (in terms of R-squared) obtained after fitting a log-linear model to each airport data. The second row (b) shows examples in which we could fit a perfect log-linear regression model and thus compute the recovery rate (LGW and CDG), and an example in  which we could not fit a log-linear regression model (WAW), which suggests that this airport does not present a clear recovery pattern. Finally, the third row (c) shows the time series for these three examples as well as the baseline and detected breaking point.}
    \label{fig:exemple_recovery_rates}
\end{figure}

Figure~\ref{fig:exemple_recovery_rates}(a) shows the performance of the fitted models in terms of R-squared, from which we could observe values higher than $0.8$ for several airports. Also, Figures~\ref{fig:exemple_recovery_rates}(b)~and~(c) illustrate airports with positive (first two columns) and negative recovery rate (last column) and their respective time series from Jan 1st, 2020 to July 30th, 2020. Only two airports presented a negative recovery rate in the considered period (MAN and WAW), meaning that their activities was still decreasing after the breaking point.

\begin{figure}
    \centering
    \includegraphics[width=0.49\textwidth]{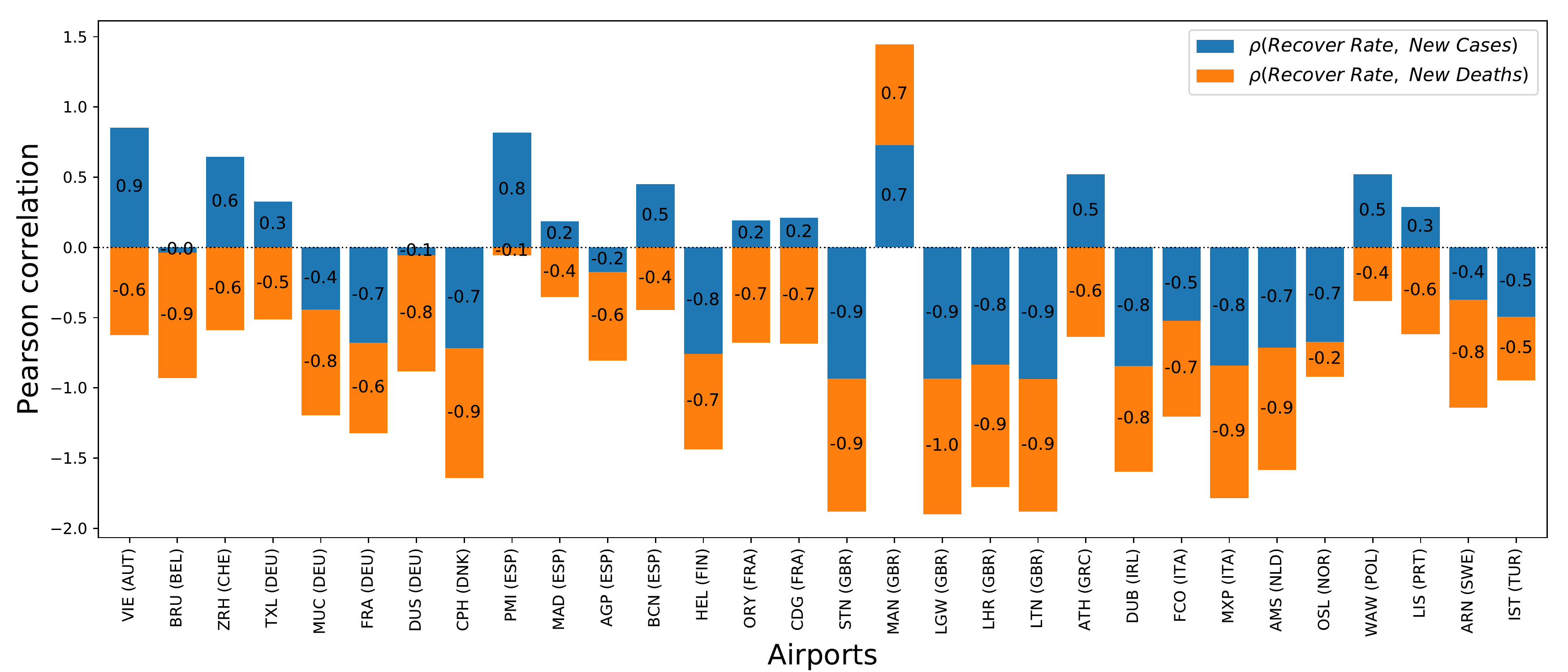}
    \caption{Correlation analysis between the recovery rates and the total number of cases and deaths caused by the COVID-19 disease.}
    \label{fig:correlation}
\end{figure}

Finally, we present an analysis of the COVID-19 situation and the public's politics for border opening by correlating the recovery rates and the 14-day moving average of the daily number of new cases and deaths, as illustrated in Figure~\ref{fig:correlation}. This study considered the official number of cases and deaths reported until July 30th, 2020~\cite{Hasell2020SciData,XU2020Lancet}. When we correlated the recovery rate and the daily number of new cases of COVID-19, we observed a strong positive correlation between both variables for VIE, PMI and ZRH. This suggests that these airports restarted their activities while the number of new cases was still increasing. Conversely, we observed a strong negative correlation between the recovery rate and the daily new cases of COVID-19 for several airports such as MUC and DUS airports from German, and the SNT, LGW, LHR, and LTN airports from Great Britain.

Of course, the positive correlations found in our analysis do not imply the causation of the rise in the total number of cases and deaths. However, we can surely state that the indicator of activity presented in this work can reveal which airports and countries demand more attention from the authorities. This because airports that are expanding their activities while the total number of cases and deaths caused by the COVID-19 disease is increasing can negatively impact the pandemic's trajectory.

\section{Final remarks}

\begin{figure}[h]
    \centering
    \includegraphics[width=0.5\textwidth]{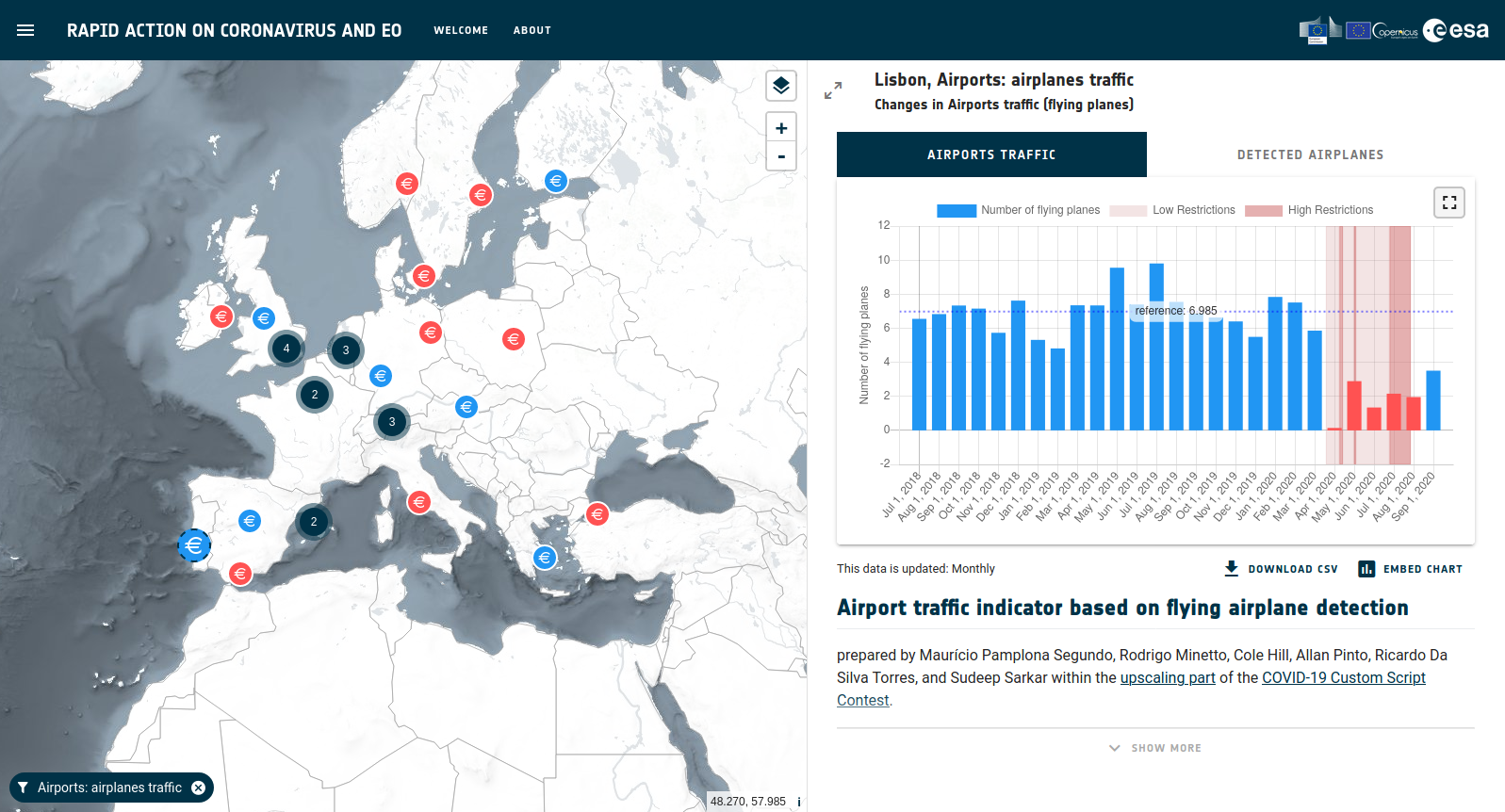}
    \vspace{1pt}

    \includegraphics[width=0.15\textwidth]{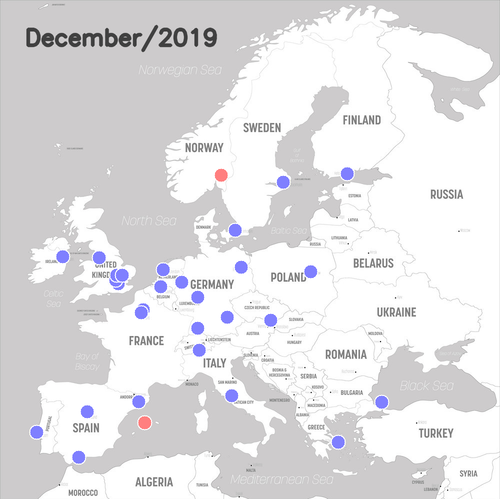} \hspace{1pt}
    \includegraphics[width=0.15\textwidth]{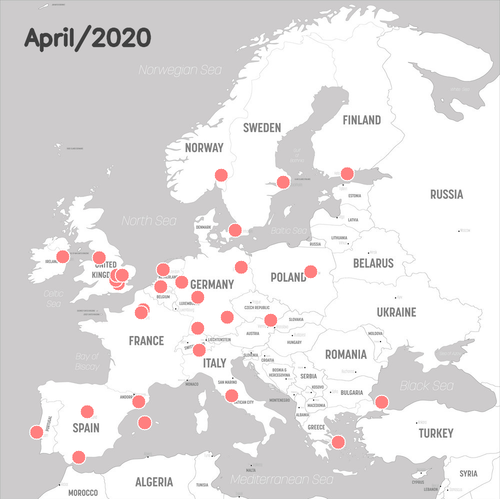} \hspace{1pt}
    \includegraphics[width=0.15\textwidth]{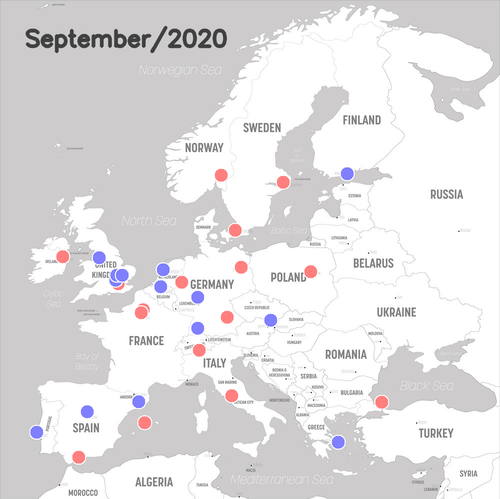}
    
    \caption{RACE observation dashboard. The map shows the location of the 30 airports being monitored. Red and blue circles in the map indicate low and normal traffic levels, respectively, and dark blue circles show the numbers of airports in that region. The bar chart in the right shows the results of our method for a specific airport, which comprises the time series of the number of flying airplanes. Finally, the three maps in the bottom shows the traffic levels for three time periods: before the COVID-19 pandemic (Dec/2019); early period after the pandemic, with mobility restrictions (Apr/2020); and after a gradual reopening (Sep/2020).}
    \label{fig:race-dashboard}
\end{figure}

The measurements of human activity are nowadays an essential task for planning actions to fight against huge outbreaks that impact human daily activities as the COVID-19 pandemic. In this context, the proposed approach for measuring airport activities can serve society as a valuable and independent indicator of human activity without any political biases. Our solution now integrates the \emph{Rapid Action Coronavirus Earth} (RACE) observation dashboard\cite{RACEtraffic}, a platform from the European Space Agency (ESA) that uses Earth observation satellite data and artificial intelligence to measure the impact of the coronavirus lockdown and to monitor post-lockdown recovery, as illustrated in Figure~\ref{fig:race-dashboard}.

Our detector was able to locate most airplanes appearing on satellite images with minimal false detections. On average, we get less than one false detection per month in each airport. Furthermore, our time series analysis can be used to identify abnormal behaviors in air traffic and correlate changes in the number of airplanes with COVID-19 statistics. Decision-makers can use this information to substantiate border control and lockdown measures.

Thanks to our semi-automatic annotation strategy, this approach can be straightforwardly adapted to detect other objects, such as ships~\cite{Heiselberg2019} and transportation trucks~\cite{RACEtrucks}, and use this outcome to devise other informative indicators.

\section*{Acknowledgments}

Part of the equipment used in this project are supported by a grant (CNS-1513126) from the USA National Science Foundation. We gratefully acknowledge the support of NVIDIA Corporation with the donation of the Titan Xp GPU used for this research. Funding from the University of South Florida for the Institute for Artificial Intelligence (AI+X) is also acknowledged. The authors would like to thank also the research Brazilian agencies CNPq, CAPES and FAPESP. 

\bibliographystyle{IEEEtran}
\bibliography{bibliography}

\end{document}